%% file: main.tex
\providecommand{\withgemma}{1}
\documentclass{article}
\PassOptionsToPackage{table}{xcolor}
\usepackage{iclr2027_conference,times}
\iclrfinalcopy
\usepackage[T1]{fontenc}
\usepackage{hyperref,url,booktabs,amsmath,amssymb,amsthm,graphicx,microtype,xcolor}
\usepackage{xurl}
\usepackage{enumitem,multirow,algorithm,algorithmic,tabularx,array}
\usepackage{caption}
\usepackage{listings,placeins,wrapfig,afterpage}

\definecolor{tzblue}{HTML}{4F7DB3}
\definecolor{tzsalmon}{HTML}{D97C74}
\definecolor{tzpurple}{HTML}{9A8BC9}
\definecolor{tzgreen}{HTML}{5E9E7E}
\definecolor{tzgrey}{HTML}{6B6B6B}
\definecolor{tzrow}{HTML}{EDF2F8}
\definecolor{tzup}{HTML}{278579}
\definecolor{tzdown}{HTML}{9C4A4A}
\definecolor{tzlink}{HTML}{28658A}

\newcommand{\best}[1]{\textbf{#1}}
\newcommand{\second}[1]{\underline{#1}}

\newcommand{\method}{Tempo-Zero}
\newif\ifgemma
\ifnum\withgemma=1 \gemmatrue\fi
\newcommand{\library}{\mathcal{P}}

\newcommand{\bank}{\mathcal{B}}

\newcommand{\rS}{r^{\mathrm{S}}}
\newcommand{\rQ}{r^{\mathrm{Q}}}
\newcommand{\gt}{y^{\star}}
\newcommand{\gtint}{e^{\star}}

\newcommand{\ourrow}{\rowcolor{tzrow}}
\newcommand{\up}[1]{{\fontsize{5.6}{6}\selectfont\textcolor{tzup}{\raisebox{0.15ex}{\scalebox{0.7}{\fontsize{5}{6}\selectfont$\uparrow$}}\,#1}}}
\newcommand{\down}[1]{{\fontsize{5.6}{6}\selectfont\textcolor{tzdown}{\raisebox{0.15ex}{\scalebox{0.7}{\fontsize{5}{6}\selectfont$\downarrow$}}\,#1}}}
\newcommand{\withdelta}[2]{\vtop{\baselineskip=6.2pt\lineskiplimit=-\maxdimen\halign{\hfil##\hfil\cr#1\cr#2\cr}}}

\newcommand{\E}{\mathbb{E}}
\newcommand{\ind}{\mathbf{1}}

\newenvironment{sequation}{\fontsize{9}{11}\selectfont\setlength{\abovedisplayskip}{5pt plus 2pt minus 2pt}\setlength{\belowdisplayskip}{5pt plus 2pt minus 2pt}\begin{equation}}{\end{equation}\ignorespacesafterend}
\DeclareMathOperator*{\mean}{mean}


\renewcommand{\arraystretch}{1.0}
\lstdefinestyle{prompt}{basicstyle=\fontsize{7.6}{9.2}\selectfont\ttfamily,breaklines=true,
 columns=fullflexible,keepspaces=true,showstringspaces=false,frame=single,
 rulecolor=\color{tzrow},backgroundcolor=\color{tzrow},framesep=5pt,xleftmargin=5.4pt,xrightmargin=5.4pt,
 literate={–}{{\textendash}}1 {—}{{\textemdash}}1 {≤}{{$\leq$}}1 {→}{{$\rightarrow$}}1}
\hypersetup{
  hidelinks,
  pdftitle={Tempo-Zero: Recursive Self-Play in Executable Worlds for Video Reasoning},
  pdfauthor={Jingmin Zhu, Yanyan Zou, Xiuyu Li, James Bailey, Dinh Phung, Tao Yu, Yunyun Hou, Jun Liu, Jun Zhou, Sirui Wang, Nan Duan, Jiachen Liu, Qiuhong Ke}
}

\title{\method{}: Recursive Self-Play\\in Executable Worlds for Video Reasoning}
\author{\textbf{Jingmin Zhu$^{1}$, Yanyan Zou$^{2,\dagger}$, Xiuyu Li$^{1}$, James Bailey$^{1}$, Dinh Phung$^{1}$} \\
\textbf{Tao Yu$^{2,3,\ddagger}$, Yunyun Hou$^{2,4,\ddagger}$, Jun Liu$^{5}$, Jun Zhou$^{6}$} \\
\textbf{Sirui Wang$^{2}$, Nan Duan$^{2}$, Jiachen Liu$^{2,\dagger}$, Qiuhong Ke$^{1,\dagger}$} \\
\normalfont $^{1}$Monash University, $^{2}$JD.com \\
\normalfont $^{3}$Institute of Automation, Chinese Academy of Sciences \\
\normalfont $^{4}$Communication University of China, $^{5}$Lancaster University, $^{6}$Griffith University \\
\normalfont $^{\dagger}$Corresponding authors, $^{\ddagger}$Internship}

\begin{document}
\raggedbottom
\maketitle
\lhead{Preprint}
\input{sections/00-abstract}
\noindent\textbf{Project link:} \url{https://alchemist-jo.github.io/tempo-zero/}

\input{sections/01-introduction}
\input{sections/02-related-work}
\input{sections/03-method}
\input{sections/05-experiments}
\input{sections/06-analysis}
\input{sections/07-conclusion}
\clearpage
\input{sections/08-statements}

\clearpage
\bibliography{references}
\bibliographystyle{iclr2027_conference}

\clearpage
\appendix
\setlength{\intextsep}{12pt plus 2pt minus 2pt}
\setlength{\textfloatsep}{14pt plus 2pt minus 2pt}
\setlength{\floatsep}{12pt plus 2pt minus 2pt}
\setcounter{topnumber}{2}
\setcounter{totalnumber}{3}
\raggedbottom
\makeatletter
\setlength{\@fptop}{0pt}
\setlength{\@fpsep}{16pt}
\setlength{\@fpbot}{0pt plus 1fil}
\makeatother
\input{sections/A-appendix}
\end{document}

%% file: sections/00-abstract.tex
\begin{abstract}
Video reasoning requires understanding how events unfold over time, yet videos with accurate temporal annotations are expensive to collect. Self-play offers an alternative by letting a model generate and solve its own tasks without manual annotation. However, existing video self-play methods train on fixed sets of recorded videos, so they cannot alter the underlying event dynamics and rely on noisy pseudo-labels {derived from the model's own outputs, such as majority voting,} that risk reinforcing shared model errors. To overcome these limitations, we introduce \method{}, a self-improvement framework that closes a loop of \emph{executable environments}, \emph{execution-verified self-play}, and \emph{environment evolution}, where executing a program renders a video with the ground-truth option and interval of every question. A single policy plays both Questioner and Solver, proposing questions and answering them with timestamped evidence lines, and is trained on 
execution-grounded rewards instead of model consensus. The Solver's performance steers a curriculum over tasks and difficulty levels, and the policy selects rule changes that enrich the event dynamics of later rounds. Despite being trained solely on simple programs, without supervised fine-tuning or external video data which prior methods typically require, \method{} outperforms the base models and recent self-play baselines on average across twelve real-world video benchmarks at different model scales, with particularly strong gains on temporal grounding tasks. Code will be made publicly available.
\end{abstract}

%% file: sections/01-introduction.tex
\section{Introduction}
\label{sec:intro}

Video understanding and reasoning require models to interpret dynamic events, track temporal dependencies, and perform multi-step inference over time~\citep{chen2025longvideoreason,rasheed2025videomathqa,hu2026videommmu}. While supervised fine-tuning and reinforcement learning for vision-language models have improved these capabilities~\citep{bai2025qwen3vl,wang2025videorft,chen2025longvideoreason}, progress remains constrained by the cost of collecting diverse videos, designing multi-step tasks, and verifying temporal annotations~\citep{rasheed2025videomathqa,huang2024lita}. To reduce dependence on human annotation, self-improvement has emerged as a promising alternative paradigm, enabling models to construct their own training tasks and improve through self-generated feedback~\citep{zou2026datacentric,zhao2025absolutezero,huang2025rzero,liu2025spice}.

\begin{wrapfigure}{R}{0.37\textwidth}
\vspace{-8pt}
\centering
\captionsetup{font=footnotesize,justification=raggedright,singlelinecheck=false}
\includegraphics[width=\linewidth]{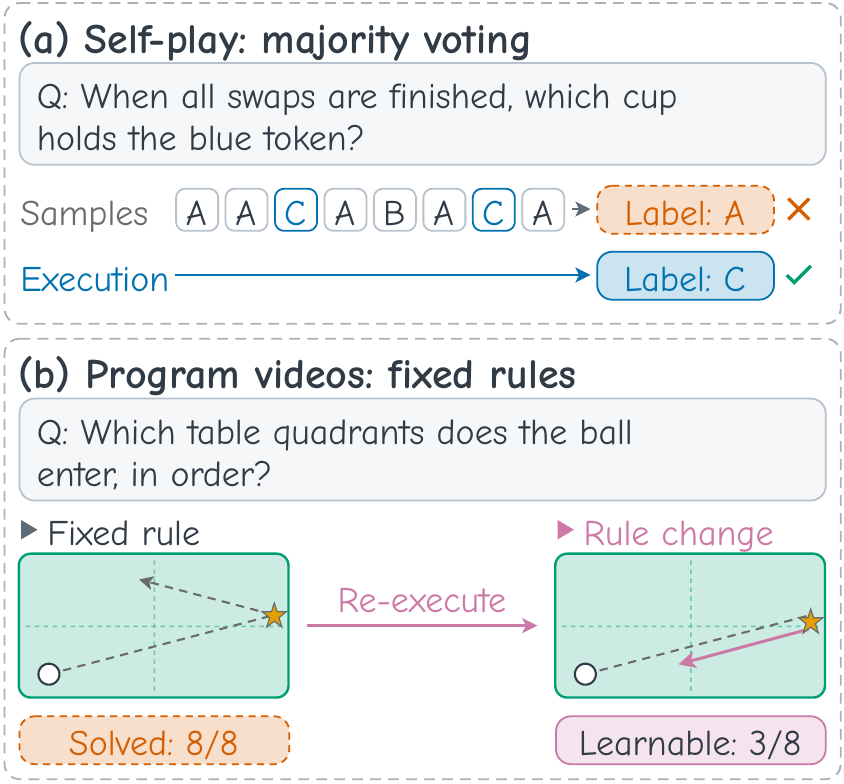}
\caption{\textbf{Limitations of video self-play and program-generated video.}
(a)~On a shell-game video, majority voting over eight sampled answers selects a wrong label over the two correct ones, while execution gives the correct label.
(b)~Under a fixed rule, all Solver rollouts are correct; changing the cushion rule and re-executing the program yields a new path with exact labels.}
\label{fig:challenges}
\vspace{-5pt}
\end{wrapfigure}

Self-play  provides a practical framework for self-improvement: a model proposes tasks, solves them, and updates its policy 
{based on the outcomes of its attempts}~\citep{silver2017alphagozero,chen2024spin,bailey2026sgs}. In text, it has improved coding and mathematical reasoning, with answers 
{verified either through programmatic execution}~\citep{zhao2025absolutezero}, by majority voting over sampled solutions~\citep{huang2025rzero}, or against source documents~\citep{liu2025spice}. Beyond text, image self-play 
{derives answer labels either through majority voting}~\citep{he2026visplay,wang2026vzero} or 
{from the outcomes of visual games executed on} rendered and edited images~\citep{wang2026visionzero}, yet static images lack temporal dependencies and continuously evolving states. Extending this recipe to video, frameworks such as Video-Zero~\citep{zhang2026videozero} and EvoVid~\citep{huang2026evovid} train on fixed collections of pre-recorded, unannotated videos, which introduces two limitations. 
First, because these videos lack verifiable ground truth, answers are labeled {from the model's own outputs, for example} by majority voting over 
{responses sampled from the same model} {(Figure~\ref{fig:challenges}a)}, which risks reinforcing systematic errors shared across samples and propagating them through training. Second, the model can vary its questions but not the events it 
{observes. Consequently,} the event dynamics and temporal complexity of its training data 
{remain fixed even as} its reasoning improves.

Executable programs offer a foundation for verifiable and controllable video self-play. In code self-play, 
{executing a model-proposed program on an input yields an automatically verifiable answer}~\citep{zhao2025absolutezero}. 
For videos, an executable program produces both the visual input and its event trace: its entities and transition rules determine the event dynamics in the rendered video, while task templates construct questions from the recorded events. The trace therefore provides each question's correct answer option and the exact time interval of the evidence needed to answer it~\citep{yi2020clevrer,jiang2026synrl}. Thus, programs control both the videos shown to the model and the ground-truth supervision used to train it, and re-executing a program under changed rules updates both.
Although program-generated videos already serve as training data, programs designed before training fix the templates, task allocation, difficulty, and event dynamics~\citep{brown2025simsv,jiang2026synrl}. The resulting training distribution cannot adapt to the learner's changing strengths and weaknesses, limiting its exposure to appropriately challenging temporal structures needed for continued improvement (Figure~\ref{fig:challenges}b). Self-play on recorded videos thus adapts only its questions and cannot verify its labels, whereas program-generated video has verifiable labels but adapts neither its questions nor its events to the learner.

Based on these observations, we introduce \method{}, which turns executable programs into an evolving source of video experiences and verifiable supervision. The framework has three components. First, \emph{executable environments} execute programs to render videos and produce their event traces. Each trace yields a set of verifiable criteria, each specifying a question, its correct answer, and the time interval of the event that determines that answer. To ensure that this supervision is available to a frame-based video model, we recompute each criterion on the sampled-frame timeline and discard it if its answer differs from the execution label. We also screen for temporal dependence by recomputing each criterion under reversed and permuted event order, removing criteria whose answers remain unchanged under both transformations. 
The retained criteria therefore describe events that the sampled frames show. Second, \emph{execution-verified self-play} uses a single shared vision-language policy  as both Questioner and Solver. The Questioner proposes questions linked to verified criteria, and the Solver answers them with timestamped evidence lines. Execution rewards the Questioner for valid questions near the Solver's current capability and rewards the Solver for the correct answer, granting interval credit only when a cited line states the correct event, rather than relying on model consensus.
Finally, \emph{environment evolution} adapts not only which tasks are sampled but also what can happen in the videos. Pass rates reallocate training across tasks and difficulty levels, and the policy selects validated rule changes, whose re-execution generates new event dynamics together with recomputed answers and evidence intervals. \method{} therefore moves beyond asking different questions about fixed videos: it creates progressively targeted video worlds whose supervision stays exact.
The loop is recursive: the policy trained in each round selects the rules that generate the videos and labels of the next round.

Our main contributions are summarized as follows:
\begin{itemize}[leftmargin=1.25em,itemsep=2pt,topsep=3pt,parsep=0pt]
\item \textbf{Executable environments for video self-play.} We formalize video task generation as program execution, where rule changes produce new event dynamics with ground-truth options and intervals, without recorded videos or manual annotations
(\S\ref{sec:environments}).
\item \textbf{Execution-verified self-play with environment evolution.} One policy plays Questioner and Solver and learns from execution-grounded 
rewards instead of majority voting, while a pass-rate curriculum and policy-selected rule changes evolve the environments with the learner (\S\ref{sec:selfplay}--\ref{sec:evolution}).
\item \textbf{Performance gains on real-world videos.} Trained solely on rendered videos, without supervised fine-tuning, \method{} improves over the base models on average \ifgemma across twelve real-world video benchmarks on four backbones from two model families\else across twelve real-world video benchmarks at three model scales\fi{} and outperforms recent self-play baselines, especially in temporal grounding (\S\ref{sec:experiments}).
\end{itemize}

%% file: sections/02-related-work.tex
\section{Related Work}
\label{sec:related}

\noindent\textbf{Self-play.}
Self-play has evolved from optimizing policies on self-generated responses to constructing tasks and evaluation signals autonomously. Early text methods contrast a model's answers with reference demonstrations~\citep{chen2024spin} or train it to score its own outputs~\citep{yuan2024selfreward}. Later frameworks split self-play into a Questioner and a Solver with feedback from code execution~\citep{zhao2025absolutezero}, majority voting~\citep{huang2025rzero}, text corpora~\citep{liu2025spice}, or a guide model~\citep{bailey2026sgs}, and self-improving agents modify their own code~\citep{zhang2026dgm,wang2026hgm} or co-evolve their evaluators~\citep{iacob2026rqgm}. In the visual domain, image self-play labels answers by majority voting~\citep{he2026visplay,wang2026vzero} or by visual games~\citep{wang2026visionzero}. Video self-play then brings time into the loop: Video-Zero adds temporal evidence to question answering~\citep{zhang2026videozero}, EvoVid rewards questions whose Solver confidence drops when frames are permuted~\citep{huang2026evovid}, and video reinforcement learning rewards temporal order~\citep{feng2025videor1,he2026videossr} or cited evidence~\citep{meng2026openo3video,liu2026videotemp}. These video methods, however, train on fixed collections of recorded videos and take their labels and intervals from annotations, model consensus, or the position of a sampled clip. \method{} removes both dependencies: it generates the videos from executable programs, so every label, including the time span of each event, is computed by execution rather than by model consensus, and its temporal checks screen these labels before any rollout.

\noindent\textbf{Programmatic environments.}
Programmatic synthesis couples simulation with deterministic supervision and turns execution traces into verifiable ground truth. For video, SynRL synthesizes videos with automated temporal supervision~\citep{jiang2026synrl}, and \mbox{SIMS-V}~\citep{brown2025simsv} and SpatialEvo~\citep{li2026spatialevo} turn 3D simulation into spatial question-answering targets. Executable code also represents worlds for physical reasoning~\citep{wang2026codeasworld} and video generation~\citep{li2026videococo}. In SynRL and \mbox{SIMS-V}, however, the rules and question templates are fixed before training, so the data cannot follow the learner. Adaptive curricula address this gap: ACCEL mutates environment parameters~\citep{parkerholder2022accel}, Prioritized Level Replay revisits levels with high learning potential~\citep{jiang2021plr}, and SPADE trains agents to construct executable environments~\citep{liu2026spade}. \method{} brings this adaptation to video self-play and closes the loop between the learner and its environments. Its pass rates steer the curriculum, the policy itself selects rule changes, and re-execution recomputes every video and label. The training distribution thus grows with the learner, and its supervision stays exact.

%% file: sections/03-method.tex
\section{Method}
\label{sec:method}

\begin{figure}[t]
\centering
\includegraphics[width=0.86\linewidth]{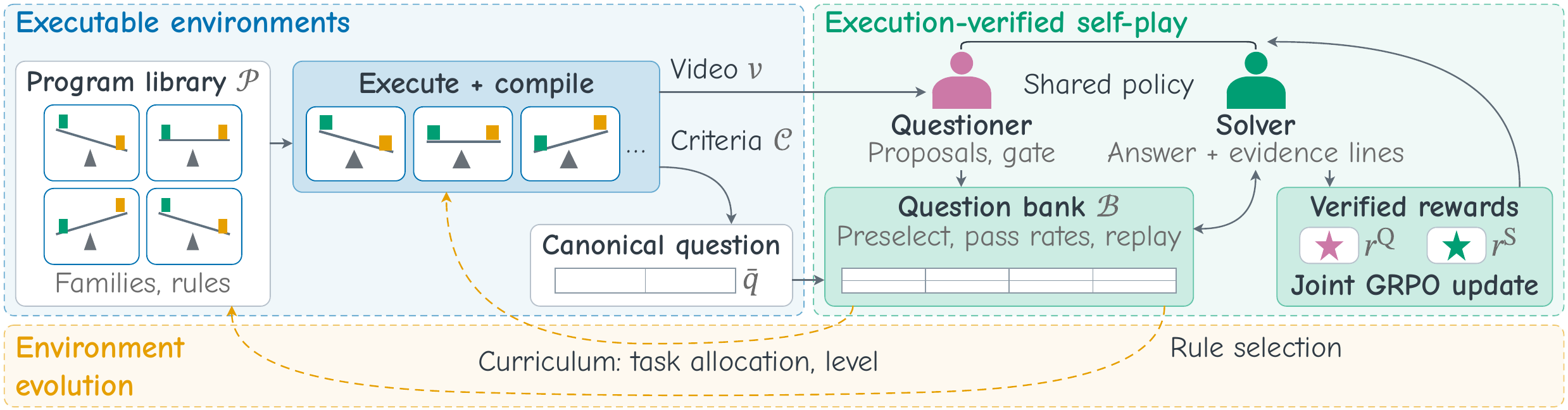}
\caption{\textbf{Overview of \method{}.}
Programs in library $\library$ execute and compile into video~$v$, criteria~$\mathcal{C}$, and canonical question~$\bar q$.
One policy proposes questions as the Questioner and answers them as the Solver; bank~$\bank$ stores measured questions for replay, and execution-grounded rewards $\rQ$ and $\rS$ drive GRPO updates.}
\label{fig:step}
\vspace{-4pt}
\end{figure}

\method{} trains a vision-language policy by self-play in an executable program library, where every label comes from program execution rather than majority voting (Figure~\ref{fig:step}).
Programs first execute into rendered videos and verification criteria (\S\ref{sec:environments}).
One policy then proposes and answers questions, and the execution trace scores both roles (\S\ref{sec:selfplay}).
Finally, a pass-rate curriculum and policy-selected rule changes evolve the environments with the learner (\S\ref{sec:evolution}).

\noindent\textbf{Problem formulation.}
We formalize video self-improvement as an interaction between two roles of a single policy $\pi_\theta$: a Questioner $\pi_\theta^{\mathrm{Q}}$ that proposes questions and a Solver $\pi_\theta^{\mathrm{S}}$ that answers them.
The two roles share parameters $\theta$ and differ only in their prompts, so visual representations learned while solving also inform question generation.
Executing a program from library $\library$ yields a synthetic video $v$ with verification criteria $\mathcal{C}$, each recording a question stem, its ground-truth option, and the time interval of the event that decides it.
With the criteria as the source of all rewards, the policy is trained to maximize
\begin{sequation}
J(\theta)=\E_{w}\,\E_{q\sim\pi_\theta^{\mathrm{Q}}(\cdot\mid w)}\,\rQ(q)\;+\;\E_{q\in\mathcal{Q},\,y\sim\pi_\theta^{\mathrm{S}}(\cdot\mid v_q,q)}\,\rS(y,q),
\label{eq:objective}
\end{sequation}
where instance $w$ is drawn by the curriculum (\S\ref{sec:evolution}), $\mathcal{Q}$ holds the questions that train the Solver, namely the compiled canonical question of each drawn instance, at most one proposal per instance, and replayed bank questions (\S\ref{sec:selfplay}), and $v_q$ is the video of question $q$.
Questioner reward $\rQ$ prioritizes valid questions near the limit of the Solver's current capability, and Solver reward $\rS$ scores the answer together with how well its cited evidence matches the execution trace.
Answered questions enter bank $\bank$ with their pass rates, the fraction of Solver rollouts that answer correctly.

\subsection{Executable Environments}
\label{sec:environments}

Executable environments compute labels with the same process that produces the video.

\begin{wrapfigure}{R}{0.36\textwidth}
\vspace{-8pt}
\centering
\captionsetup{font=footnotesize,justification=raggedright,singlelinecheck=false}
\includegraphics[width=\linewidth]{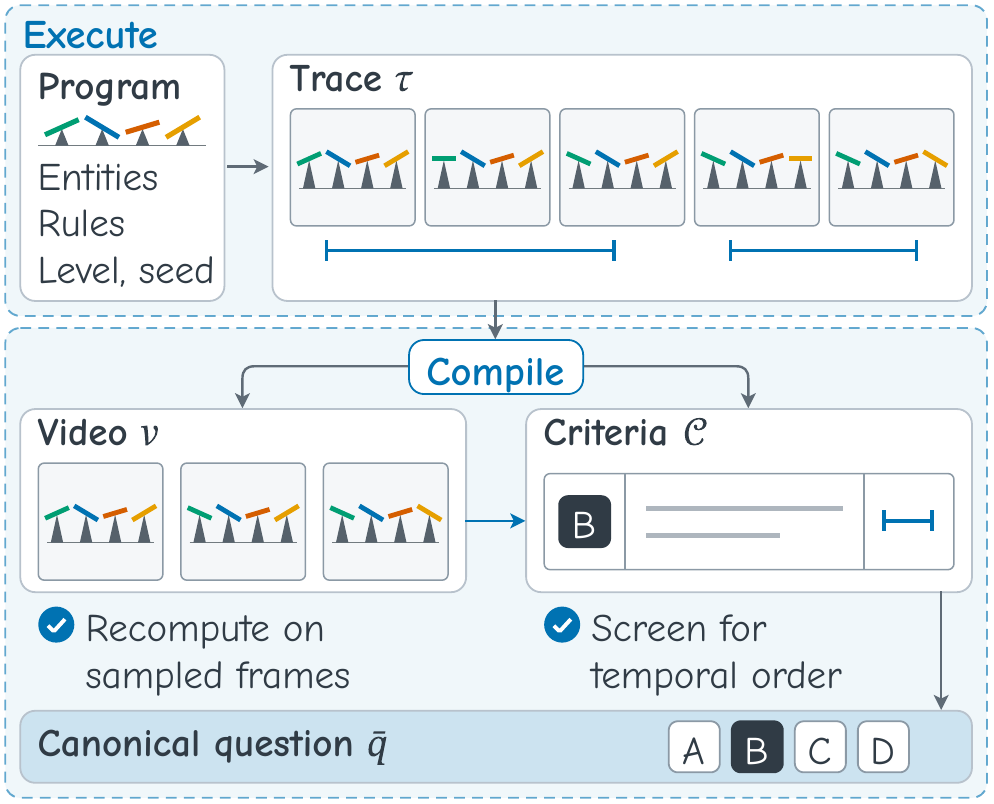}
\caption{\textbf{Environment compilation.}\label{fig:environment}
Ground truth is recomputed on the sampled frames, and criteria are screened for temporal order.
Colors identify entities, blue brackets mark ground-truth intervals, and the dark option is the ground-truth option.}
\vspace{-8pt}
\end{wrapfigure}

\noindent\textbf{Programs and instances.}
Library $\library$ groups programs into environment families $f$.
Each family defines physical entities, state-transition rules with alternative values, and the tasks it supports.
Within a family, an instance $w$ is set by three quantities: difficulty level $\ell$ scales entity counts and operation complexity, rule configuration $u_f$ selects the active transition behaviors, and playback scale $\lambda_r$ sets video duration.
Instance parameters $x$ fix the remaining randomness (Appendix~\ref{app:environments}).
Given these inputs, program $p_f$ executes into trace $\tau$, which compiles into video $v$, criteria $\mathcal{C}$, and canonical question $\bar q$ (Figure~\ref{fig:environment}):
\begin{sequation}
\tau=\mathrm{Exec}(p_f;\,\ell,u_f,x),\qquad (v,\mathcal{C},\bar q)=\mathrm{Compile}(\tau;\,a,\lambda_r).
\label{eq:execute}
\end{sequation}
Because $v$ and $\mathcal{C}$ come from the same trace, changing $\ell$ or $u_f$ changes the event dynamics and the ground truth together.
Compile writes $\bar q$ for the task $a$ allocated to the instance, using the stem and ground-truth option of the matching criterion and distractors from its domain.

\noindent\textbf{Criteria and tasks.}
Execution traces yield reliable supervision only when each question is tied to a dynamic event that the policy can see.
Each learning target is therefore formalized as criterion $c\in\mathcal{C}$.
A criterion specifies a question stem, ground-truth option $\gt$, ground-truth interval $\gtint_c$ (one or more time segments), and a distractor domain with a minimum resolution floor.
An execution label replaces model consensus only if it holds on the frames the policy sees and its answer depends on event order, so each criterion passes two programmatic checks before training.

\noindent\emph{Frame recomputation.}
Programs advance state on a fine simulation clock, while the policy observes only frames sampled at a much lower rate, so when an event starts or ends between two sampled frames, a label read from the simulation clock describes a state that no sampled frame shows.
We therefore express every temporal label on the timestamps of the sampled frames and recompute each ground-truth value from those frames.
A criterion is discarded when its value cannot be recomputed from those frames or disagrees with the simulation (Appendix~\ref{app:check_stats}).

\noindent\emph{Static-question screen.}
A question whose answer does not depend on the order of events is static.
To detect such questions, we recompute each criterion's answer on the reversed trace and on a trace with permuted events, and we exclude criteria whose answer stays the same under both transformations.
When a transformation yields no determinate answer, the screen keeps the criterion, which must still show its target in several sampled frames (Appendix~\ref{app:sampling_coords}).

After both checks, the retained criteria span the tasks in Table~\ref{tab:tasks}, such as event count, event order, and state change, grouped into seven task classes.
Each family supports a subset of these tasks, and the (family, task) pair $(f,a)$ serves as the scheduling unit of the curriculum (\S\ref{sec:evolution}).

\subsection{Execution-Verified Self-Play}
\label{sec:selfplay}

\begin{wrapfigure}{r}{0.36\textwidth}
\vspace{-8pt}
\centering
\captionsetup{font=footnotesize,justification=raggedright,singlelinecheck=false}
\includegraphics[width=\linewidth]{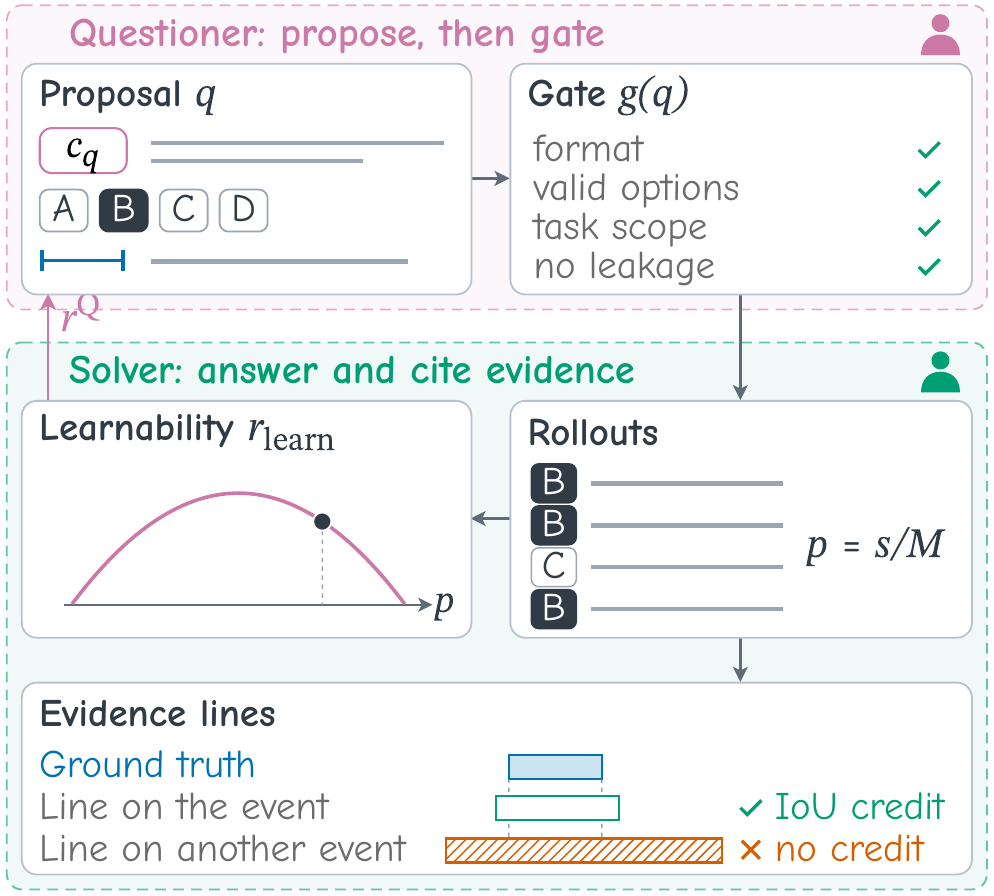}
\caption{\textbf{Execution-verified self-play.}\label{fig:selfplay-method}
The Questioner's proposal is bound to criterion~$c_q$ and passes gate~$g(q)$.
The Solver's rollouts give pass rate~$p$, which sets $r_{\mathrm{learn}}$ for the Questioner, and only an evidence line that states the criterion's event earns interval credit.}
\end{wrapfigure}

Given compiled instances, self-play needs a validity check for each proposed question and a reward for each answer.
\method{} computes both from the execution trace, so the loop involves no judge model or majority voting (Figures~\ref{fig:challenges}a and~\ref{fig:step}).
In each step, the Questioner proposes questions, a programmatic gate filters them, the Solver answers them with timestamped evidence lines, and bank $\bank$ keeps questions for replay (Figure~\ref{fig:selfplay-method}).

\noindent\textbf{Question proposals.}
Each compiled instance already carries its canonical question $\bar q$, which the Solver answers whether or not any proposal passes the gate.
The Questioner asks about the same verified events in new ways: it selects one of the criteria and writes a new stem and distractors, so the policy adapts its questions to the Solver's current capability while the criterion fixes the ground truth.
Because the stem and options are free-form, they can be ambiguous or invalid, or leak the answer.
To write a proposal, the Questioner receives the video, the askable criteria in $\mathcal{C}$ without the timestamps of their events, same-task questions from the bank, and the Solver's pass rate per criterion (Appendix~\ref{app:protocols}).
Its output, a proposal $q$, names one criterion $c_q\in\mathcal{C}$ and contains a question stem, the ground-truth option $\gt$ of $c_q$, distractors, and evidence lines citing the events that the criterion depends on.
Since its input gives no timestamps, the intervals of these lines are information that the proposal adds, which novelty reward $r_{\mathrm{nov}}$ below scores.
Each proposal then passes through a programmatic gate $g(q)\in\{0,1\}$ that checks format, option validity, task scope, and answer leakage in the stem.
Option validity requires exactly one option to equal $\gt$ and every distractor to be separated from $\gt$ and from the other distractors by the resolution floor.
For accepted proposals, the environment shuffles the options uniformly at random and assigns option letters; proposals with $g=0$ are not answered.

\noindent\textbf{Question bank.}
Bank $\bank$ serves three purposes: it supplies same-task examples and pass rates to the Questioner, replay questions to the Solver, and pass-rate statistics to the curriculum (\S\ref{sec:evolution}).
Among the gated proposals, it preselects at most one per instance, favoring rare task classes and tasks, and stores it with each measured canonical question.
At capacity, eviction first removes overrepresented task classes and families, then the oldest measurements, so every task class stays represented.

\noindent\textbf{Answers and evidence.}
On the Solver side, an answer letter alone does not show whether the Solver located the events that decide it, since multiple-choice answers can often be inferred from isolated frames.
The Solver therefore cites its evidence.
It receives only the video, the question, and the options, and it answers every question with $M$ rollouts under $\pi_\theta^{\mathrm{S}}$.
Across the $M$ rollouts, pass rate $p(q)=s/M$, where $s$ rollouts are correct, measures question difficulty.
Rollouts on every gated proposal score the Questioner, and those on canonical questions, preselected proposals, and replayed questions also train the Solver.
Each rollout $y$ contains an answer letter, whose option we denote $\hat y$, and $L$ evidence lines $(I_k,d_k)$, each pairing a predicted interval $I_k$ with an event statement $d_k$.
The evidence lines are scored against the criteria $\mathcal{C}_q\subseteq\mathcal{C}$ of question $q$, namely the criteria of the events its answer rests on, or $c_q$ itself when it rests on none.
The program parses the lines with a fixed grammar and assigns them one-to-one to these criteria: line $k(c)$ is assigned to criterion $c$, and $\delta_c\in\{0,1\}$ records whether it states the event of $c$, that is, asserts the recorded value of $c$ about its subject (Appendix~\ref{app:signals}).
With $m$ matched criteria, evidence score $r_{\mathrm{evid}}$ is the statement-level $F_1$ score, with precision $m/L$ and recall $m/|\mathcal{C}_q|$.
Grounded score $r_{\mathrm{grd}}$ averages the temporal Intersection over Union (IoU) over the criteria and counts only matched lines:
\begin{sequation}
r_{\mathrm{evid}}=\frac{2m}{L+|\mathcal{C}_q|},\qquad
r_{\mathrm{grd}}=\mean_{c\in\mathcal{C}_q}\; \delta_c\,\mathrm{IoU}\bigl(I_{k(c)},\gtint_c\bigr).
\label{eq:evidence_scores}
\end{sequation}
A simpler grounding reward, $\max_k\mathrm{IoU}(I_k,\gtint_c)$, would credit a broad interval on any line, even one that describes another event (Figure~\ref{fig:selfplay-method}); $r_{\mathrm{grd}}$ instead counts overlap only on the line whose statement matches the criterion, so interval credit requires the correct event.

\noindent\textbf{Reward terms.}
Both rewards combine these terms and are computed programmatically from the execution trace, the Solver's rollouts, and the bank:
\begin{sequation}
\rS(y,q)=r_{\mathrm{ans}}+\mean\bigl(r_{\mathrm{evid}},r_{\mathrm{grd}}\bigr),\qquad
\rQ(q)=g(q)\,\mean\bigl(r_{\mathrm{learn}},r_{\mathrm{nov}}\bigr).
\label{eq:rewards}
\end{sequation}
Here $r_{\mathrm{ans}}=\ind[\hat y=\gt]$ is the answer reward, or $\mathrm{IoU}(\hat I,\gtint_{c_q})$ for interval answer $\hat I$ when the ground truth is one interval, correct when at least $\epsilon$; each term lies in $[0,1]$, so $\rS\in[0,2]$.
The Solver's answer and evidence are rewarded separately, so a rollout that cites the correct event earns evidence credit even when its final choice is wrong, while $r_{\mathrm{ans}}$ still separates correct from incorrect answers.
For the Questioner, gate $g(q)$ removes the reward of invalid proposals, and learnability reward $r_{\mathrm{learn}}$ favors questions that the Solver neither always solves nor always misses.
It is the variance of a Bernoulli variable whose success probability is the Laplace-smoothed pass rate $\tilde p$, scaled to peak at one:
\begin{sequation}
r_{\mathrm{learn}}(q)=4\,\tilde p\,(1-\tilde p),\qquad \tilde p=\frac{s+1}{M+2}.
\label{eq:learnability}
\end{sequation}
Smoothing keeps the reward positive when all rollouts are correct or all are wrong.
Finally, novelty reward $r_{\mathrm{nov}}\in[0,1]$ is $r_{\mathrm{grd}}$ computed on the proposal's own evidence lines with every interval that the question prints left unscored, so it credits only event times that the Questioner locates in the video (Appendix~\ref{app:rewards}).

\noindent\textbf{Policy update.}
Both rewards feed a single joint update of the shared parameters $\theta$.
Within step $t$, proposals, Solver rollouts, and both rewards are computed under the same snapshot $\pi_{\theta_t}$ and held fixed during the update, so each Questioner reward reflects the current Solver and passes no gradient through its pass rate.
The policy is then updated in one Group Relative Policy Optimization (GRPO) step~\citep{shao2024deepseekmath} over the groups of both roles.
A Solver group holds the $M$ rollouts of one question, and a Questioner group holds the proposals of one instance.
Within each group $\mathcal{G}$, the advantage of sample $i$ is its mean-centered reward, $\hat A_i=r_i-\bar r_{\mathcal{G}}$, where $\bar r_{\mathcal{G}}$ is the mean reward of $\mathcal{G}$.
Groups whose rewards are all equal carry no advantage and are omitted from the update.

\subsection{Environment Evolution}
\label{sec:evolution}

\begin{wrapfigure}{r}{0.36\textwidth}
\vspace{-8pt}
\centering
\captionsetup{font=footnotesize,justification=raggedright,singlelinecheck=false}
\includegraphics[width=\linewidth]{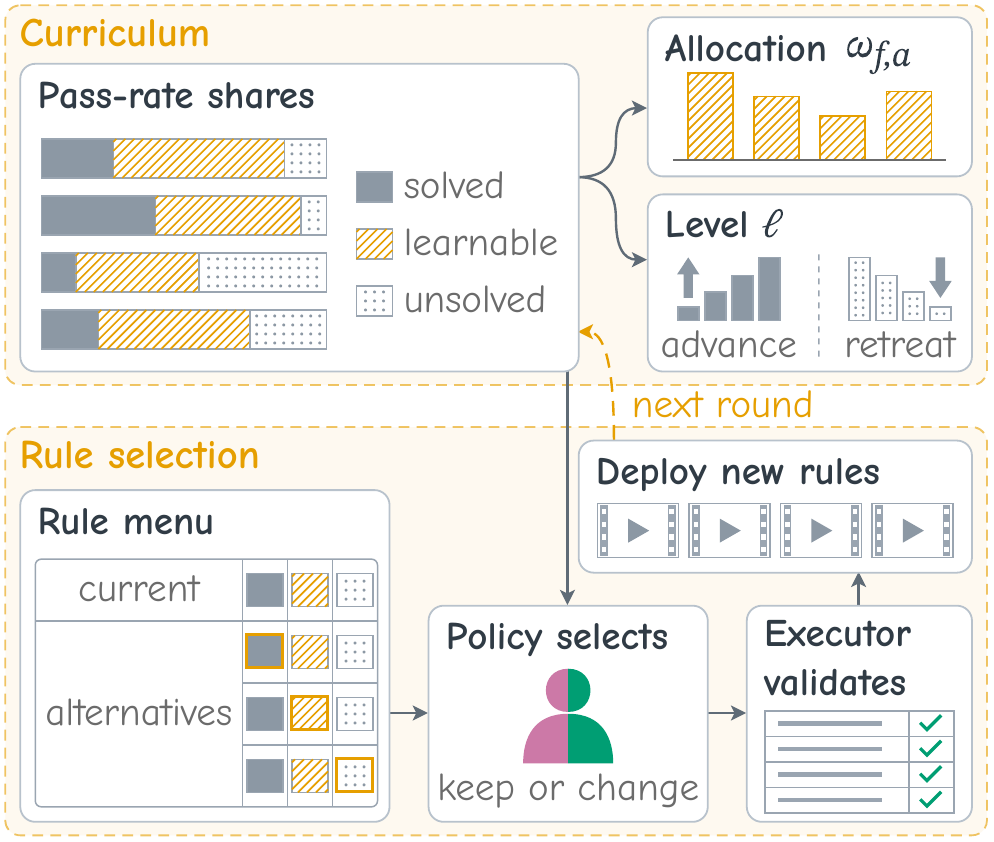}
\caption{\textbf{Environment evolution.}\label{fig:evolution-method}
Pass rates guide allocation $\omega_{f,a}$ (Eq.~\ref{eq:weight}) and set level $\ell$, and the policy selects a rule change, which the executor validates before deployment.}
\vspace{-8pt}
\end{wrapfigure}

Fixed programs freeze the task allocation, difficulty, and event dynamics of training (Figure~\ref{fig:challenges}b). As the policy improves, more Solver groups agree, so the answer reward no longer separates their rollouts.
\method{} therefore changes the environments in two ways (Figure~\ref{fig:evolution-method}): the curriculum reallocates instances across (family, task) pairs and levels by pass rate, and the policy selects rule changes that the executor validates.

\noindent\textbf{Task allocation.}
Uniform instance sampling spends updates on tasks that the policy already solves or cannot yet solve.
To avoid this, the curriculum reads exponentially decayed pass-rate statistics in bank $\bank$ after each round and fixes the allocation and levels of the next round.
Each family receives an equal share of instances, and within family $f$, pair $(f,a)$ receives
\begin{sequation}
\omega_{f,a}=\kappa/n_f+(1-\kappa)\,w_{f,a}\big/{\textstyle\sum_{a'}}w_{f,a'},
\label{eq:weight}
\end{sequation}
where coverage share $\kappa$ spreads part of the quota evenly over the $n_f$ tasks of $f$ so that every task stays measured, and the policy sets task weight $w_{f,a}$ per class, times a fixed task tilt.
The same statistics move difficulty: a pair's level $\ell$ rises when its decayed pass rate reaches $\rho_{\mathrm{adv}}$ and falls at $\rho_{\mathrm{ret}}$.

\pagebreak

\noindent\textbf{Rule selection.}
Allocation redistributes the existing environments but cannot create new event dynamics, which requires changing the rules.
Rules decide which dynamics a family can produce, and \method{} lets the policy choose how they change.
After each round except the last, the environment runtime composes a rule menu for every family.
The menu lists the current configuration and alternatives that each move one rule, or in later rounds a pair of rules, to values the family has not run.
The policy reads the family's pass rates and recorded responses under the current configuration, sets the class weights in $w_{f,a}$ of Eq.~\ref{eq:weight}, and either keeps $u_f$ or selects a menu entry, optionally, for Shell game, with operation-graph edits (Appendix~\ref{app:rule_selection}).
Before deployment, the executor checks each selected change for validity and re-executes the program to preview the resulting videos and criteria; an invalid change keeps the previous program.
Rule selection runs at inference and receives no gradient update.
Independently of rule selection, playback scale $\lambda_r$ follows a fixed schedule that lengthens videos across rounds $r$.
Algorithm~\ref{alg:tempo} in Appendix~\ref{app:algorithm} summarizes the full training loop.

%% file: sections/05-experiments.tex
\section{Experiments}
\label{sec:experiments}

We test whether execution-verified self-play improves real-video grounding, long-video understanding, video reasoning, and other architectures.

\begin{table}[t]
\caption{\textbf{Temporal grounding on real-world benchmarks.}
Metrics in \%; Avg.\ averages the mIoU of the first three benchmarks and GQA@0.3.
Bold marks the best per column within each backbone, and underline marks the second best at 4B and 8B, where all methods are evaluated; \ifgemma arrows give the change over the base model, the only comparison at 30B-A3B and Gemma.\else arrows give the change of \method{} over the base model, the only comparison at 30B-A3B.\fi}
\label{tab:grounding}
\centering\scriptsize\setlength{\aboverulesep}{0.25ex}\setlength{\belowrulesep}{0.45ex}\setlength{\tabcolsep}{\ifgemma 1.0pt\else 1.24pt\fi}
\renewcommand{\arraystretch}{1.0}
\begin{tabular}{c@{\hspace{4pt}}l*{15}{c}}
\toprule
& \multirow{2}{*}{\textbf{Method}}
& \multicolumn{4}{c}{\textbf{ANet-RTL}}
& \multicolumn{4}{c}{\textbf{ActivityNet Captions}}
& \multicolumn{4}{c}{\textbf{Charades-STA}}
& \multicolumn{2}{c}{\textbf{NExT-GQA}} & \multirow{2}{*}{\textbf{Avg.}} \\
\cmidrule(lr){3-6} \cmidrule(lr){7-10} \cmidrule(lr){11-14} \cmidrule(lr){15-16}
&& mIoU & R@0.3 & R@0.5 & R@0.7
& mIoU & R@0.3 & R@0.5 & R@0.7
& mIoU & R@0.3 & R@0.5 & R@0.7
& GQA@0.3 & GQA@0.5 & \\
\midrule
& Base model & 16.86 & 22.70 & 15.72 & 8.73 & 22.86 & 30.94 & 19.04 & 11.29 & 37.52 & 58.90 & 36.12 & 16.23 & 23.09 & 15.40 & 25.08 \\
& VisPlay & 16.58 & 23.14 & 13.97 & 8.73 & 22.89 & 30.98 & 19.12 & 11.31 & 37.59 & 58.82 & 36.59 & 16.42 & 23.14 & 15.33 & 25.05 \\
& V-Zero & 15.85 & 20.96 & 13.97 & 8.30 & 22.81 & 30.91 & 18.92 & 11.31 & 37.10 & 58.33 & 35.24 & 15.62 & 22.82 & 15.22 & 24.64 \\
& Video-Zero & \second{29.75} & \second{44.98} & \second{28.82} & \best{13.54} & \second{32.12} & \second{44.76} & \second{27.67} & \second{16.24} & \second{41.85} & \second{65.11} & \second{40.81} & \best{20.03} & \second{38.29} & \second{23.00} & \second{35.50} \\
\ourrow \multirow{-5}{*}{\textit{4B}} & \method{} & \withdelta{\best{31.34}}{\up{14.48}} & \withdelta{\best{48.03}}{\up{25.33}} & \withdelta{\best{31.00}}{\up{15.28}} & \withdelta{\second{13.10}}{\up{4.37}} & \withdelta{\best{34.98}}{\up{12.12}} & \withdelta{\best{49.35}}{\up{18.41}} & \withdelta{\best{31.04}}{\up{12.00}} & \withdelta{\best{17.26}}{\up{5.97}} & \withdelta{\best{42.53}}{\up{5.01}} & \withdelta{\best{66.29}}{\up{7.39}} & \withdelta{\best{41.26}}{\up{5.14}} & \withdelta{\second{19.49}}{\up{3.26}} & \withdelta{\best{39.17}}{\up{16.08}} & \withdelta{\best{24.50}}{\up{9.10}} & \withdelta{\best{37.00}}{\up{11.92}} \\
\cmidrule(lr){1-17}
& Base model & 21.95 & 30.57 & 21.83 & 10.04 & 26.56 & 36.59 & 22.63 & 12.85 & 36.70 & 59.11 & 33.87 & 13.49 & 35.93 & 21.83 & 30.28 \\
& VisPlay & 21.83 & 32.31 & 20.09 & 10.04 & 26.74 & 36.94 & 22.99 & 12.93 & 36.65 & 59.03 & 33.92 & 13.47 & 35.87 & 22.19 & 30.27 \\
& V-Zero & 22.77 & 32.31 & 20.52 & 9.17 & 26.69 & 36.79 & 23.05 & 13.00 & 36.43 & 58.63 & 33.66 & 13.41 & 35.73 & 22.11 & 30.40 \\
& Video-Zero & \second{26.19} & \second{39.30} & \second{24.45} & \second{12.23} & \second{29.87} & \second{41.77} & \second{26.23} & \second{14.66} & \second{37.34} & \second{60.16} & \second{35.54} & \second{13.58} & \second{39.60} & \second{24.26} & \second{33.25} \\
\ourrow \multirow{-5}{*}{\textit{8B}} & \method{} & \withdelta{\best{31.22}}{\up{9.27}} & \withdelta{\best{46.72}}{\up{16.15}} & \withdelta{\best{28.38}}{\up{6.55}} & \withdelta{\best{13.54}}{\up{3.50}} & \withdelta{\best{34.16}}{\up{7.60}} & \withdelta{\best{49.20}}{\up{12.61}} & \withdelta{\best{30.86}}{\up{8.23}} & \withdelta{\best{16.26}}{\up{3.41}} & \withdelta{\best{46.10}}{\up{9.40}} & \withdelta{\best{72.26}}{\up{13.15}} & \withdelta{\best{49.81}}{\up{15.94}} & \withdelta{\best{23.06}}{\up{9.57}} & \withdelta{\best{40.39}}{\up{4.46}} & \withdelta{\best{25.14}}{\up{3.31}} & \withdelta{\best{37.97}}{\up{7.69}} \\
\cmidrule(lr){1-17}
& Base model & 23.45 & 32.75 & 23.14 & 13.10 & 27.90 & 37.65 & 23.98 & 14.39 & 47.34 & 68.44 & 51.61 & \best{30.13} & 38.83 & 23.63 & 34.38 \\
\ourrow \multirow{-2}{*}{\shortstack{\textit{30B-}\\\textit{A3B}}} & \method{} & \withdelta{\best{34.23}}{\up{10.78}} & \withdelta{\best{49.78}}{\up{17.03}} & \withdelta{\best{33.62}}{\up{10.48}} & \withdelta{\best{17.90}}{\up{4.80}} & \withdelta{\best{36.35}}{\up{8.45}} & \withdelta{\best{52.42}}{\up{14.77}} & \withdelta{\best{32.95}}{\up{8.97}} & \withdelta{\best{17.38}}{\up{2.99}} & \withdelta{\best{48.77}}{\up{1.43}} & \withdelta{\best{77.15}}{\up{8.71}} & \withdelta{\best{52.96}}{\up{1.35}} & \withdelta{22.37}{\down{7.76}} & \withdelta{\best{45.15}}{\up{6.32}} & \withdelta{\best{27.88}}{\up{4.25}} & \withdelta{\best{41.12}}{\up{6.74}} \\
\ifgemma\cmidrule(lr){1-17}
& Base model & 23.97 & 34.93 & 22.27 & 10.04 & 18.64 & 24.93 & 14.65 & 7.05 & 19.08 & 27.39 & 13.04 & 6.05 & 21.72 & 10.99 & 20.85 \\
\ourrow \multirow{-2}{*}{\textit{Gemma}} & \method{} & \withdelta{\best{30.52}}{\up{6.55}} & \withdelta{\best{41.48}}{\up{6.55}} & \withdelta{\best{27.51}}{\up{5.24}} & \withdelta{\best{15.72}}{\up{5.68}} & \withdelta{\best{33.53}}{\up{14.89}} & \withdelta{\best{47.52}}{\up{22.59}} & \withdelta{\best{29.94}}{\up{15.29}} & \withdelta{\best{15.99}}{\up{8.94}} & \withdelta{\best{32.01}}{\up{12.93}} & \withdelta{\best{50.97}}{\up{23.58}} & \withdelta{\best{24.30}}{\up{11.26}} & \withdelta{\best{9.52}}{\up{3.47}} & \withdelta{\best{26.40}}{\up{4.68}} & \withdelta{\best{14.64}}{\up{3.65}} & \withdelta{\best{30.61}}{\up{9.76}} \\
\fi
\bottomrule
\end{tabular}
\end{table}

\subsection{Experimental Setup}
\label{sec:setup}

\noindent\textbf{Models and training.}
We train Qwen3-VL-4B-Instruct and Qwen3-VL-8B-Instruct~\citep{bai2025qwen3vl}, each serving as both Questioner and Solver, with LoRA adapters for five rounds on videos rendered by the executable environments alone, without external videos, human annotations, or supervised fine-tuning (Appendix~\ref{app:protocol}).
Baselines are the instruction-tuned base model and three self-improvement methods: VisPlay~\citep{he2026visplay} and V-Zero~\citep{wang2026vzero}, which train on images, and Video-Zero~\citep{zhang2026videozero}, which trains on recorded videos; all results follow the evaluation protocol of~\citet{zhang2026videozero}.

\noindent\textbf{Benchmarks.}
The final checkpoint is evaluated zero-shot on twelve real-world benchmarks in three groups.
Temporal grounding uses ANet-RTL~\citep{huang2024lita}, ActivityNet Captions~\citep{krishna2017activitynet}, \mbox{Charades-STA}~\citep{gao2017tall}, and NExT-GQA~\citep{xiao2024nextgqa}, scored by mean IoU (mIoU), recall at IoU thresholds (R@0.3/0.5/0.7), and grounded QA accuracy (GQA@0.3/0.5).
Long-video understanding uses LongVideoBench~\citep{wu2024longvideobench}, MLVU~\citep{zhou2025mlvu}, Video-MME-Long~\citep{fu2025videomme}, and LSDBench~\citep{qu2025lsdbench}, and video reasoning uses LongVideoReason~\citep{chen2025longvideoreason}, MMVU~\citep{zhao2025mmvu}, VideoMathQA~\citep{rasheed2025videomathqa}, and VideoMMMU~\citep{hu2026videommmu}, both scored by accuracy (Appendix~\ref{app:panels}).

\subsection{Main Results}
\label{sec:main-results}

\begin{table}[t]
\caption{\textbf{Long-video understanding and video reasoning.}
Official accuracy in \%; marks as in Table~\ref{tab:grounding}.}
\label{tab:qa}
\centering\small\setlength{\aboverulesep}{0.25ex}\setlength{\belowrulesep}{0.45ex}\setlength{\tabcolsep}{1.0pt}
\renewcommand{\arraystretch}{1.0}
\resizebox{\linewidth}{!}{%
\begin{tabular}{c@{\hspace{4pt}}l*{9}{c}}
\toprule
& \multirow{2}{*}{\textbf{Method}}
& \multicolumn{4}{c}{\textbf{Long-Video Understanding}}
& \multicolumn{4}{c}{\textbf{Video Reasoning}} & \multirow{2}{*}{\textbf{Avg.}} \\
\cmidrule(lr){3-6}\cmidrule(lr){7-10}
&& \shortstack{LongVideo\\Bench} & MLVU & \shortstack{Video-MME\\Long} & LSDBench & \shortstack{LongVideo\\Reason} & MMVU & \shortstack{VideoMath\\QA} & VideoMMMU & \\
\midrule
& Base model & 57.70 & 62.60 & 51.60 & 54.68 & 69.90 & 64.32 & 19.05 & 57.56 & 54.68 \\
& VisPlay & 59.20 & \second{63.50} & 53.20 & 54.83 & 69.60 & 63.36 & 18.33 & 57.44 & 54.93 \\
& V-Zero & \second{59.50} & \second{63.50} & 51.90 & 54.68 & 68.20 & 64.48 & 20.95 & 58.33 & 55.19 \\
& Video-Zero & 59.10 & 63.40 & \best{53.60} & \best{56.52} & \second{70.30} & \second{64.96} & \second{21.67} & \second{61.56} & \second{56.39} \\
\ourrow \multirow{-5}{*}{\textit{4B}} & \method{} & \withdelta{\best{61.00}}{\up{3.30}} & \withdelta{\best{67.36}}{\up{4.76}} & \withdelta{\second{53.30}}{\up{1.70}} & \withdelta{\second{56.06}}{\up{1.38}} & \withdelta{\best{72.40}}{\up{2.50}} & \withdelta{\best{66.08}}{\up{1.76}} & \withdelta{\best{22.38}}{\up{3.33}} & \withdelta{\best{63.11}}{\up{5.55}} & \withdelta{\best{57.71}}{\up{3.03}} \\
\cmidrule(lr){1-11}
& Base model & 59.80 & 64.40 & 55.70 & \second{57.59} & 68.60 & 68.32 & 23.81 & 63.78 & 57.75 \\
& VisPlay & \second{60.50} & \second{64.90} & 54.90 & 57.44 & 68.10 & 67.04 & 21.43 & 62.67 & 57.12 \\
& V-Zero & 60.10 & 64.20 & 56.00 & 56.90 & 69.20 & \best{68.80} & 22.14 & 63.22 & 57.57 \\
& Video-Zero & 60.40 & 64.60 & \second{56.60} & 56.75 & \second{70.70} & 68.00 & \second{25.48} & \second{64.22} & \second{58.34} \\
\ourrow \multirow{-5}{*}{\textit{8B}} & \method{} & \withdelta{\best{61.00}}{\up{1.20}} & \withdelta{\best{68.51}}{\up{4.11}} & \withdelta{\best{56.90}}{\up{1.20}} & \withdelta{\best{58.59}}{\up{1.00}} & \withdelta{\best{73.00}}{\up{4.40}} & \withdelta{\second{68.48}}{\up{0.16}} & \withdelta{\best{27.38}}{\up{3.57}} & \withdelta{\best{64.44}}{\up{0.66}} & \withdelta{\best{59.79}}{\up{2.04}} \\
\cmidrule(lr){1-11}
& Base model & 65.37 & 70.26 & 58.33 & 62.50 & 71.40 & 72.16 & 26.90 & 65.11 & 61.50 \\
\ourrow \multirow{-2}{*}{\shortstack{\textit{30B-}\\\textit{A3B}}} & \method{} & \withdelta{\best{66.70}}{\up{1.33}} & \withdelta{\best{71.34}}{\up{1.08}} & \withdelta{\best{60.56}}{\up{2.23}} & \withdelta{\best{63.96}}{\up{1.46}} & \withdelta{\best{75.60}}{\up{4.20}} & \withdelta{\best{73.60}}{\up{1.44}} & \withdelta{\best{31.90}}{\up{5.00}} & \withdelta{\best{67.44}}{\up{2.33}} & \withdelta{\best{63.89}}{\up{2.39}} \\
\ifgemma\cmidrule(lr){1-11}
& Base model & 50.71 & 54.51 & 46.44 & 47.09 & 70.60 & 65.28 & \best{30.00} & 65.89 & 53.81 \\
\ourrow \multirow{-2}{*}{\textit{Gemma}} & \method{} & \withdelta{\best{53.40}}{\up{2.69}} & \withdelta{\best{54.62}}{\up{0.11}} & \withdelta{\best{48.00}}{\up{1.56}} & \withdelta{\best{55.14}}{\up{8.05}} & \withdelta{\best{71.10}}{\up{0.50}} & \withdelta{\best{67.04}}{\up{1.76}} & \withdelta{29.29}{\down{0.71}} & \withdelta{\best{66.11}}{\up{0.22}} & \withdelta{\best{55.59}}{\up{1.78}} \\
\fi
\bottomrule
\end{tabular}}%
\vspace{-4pt}
\end{table}

\noindent\textbf{Temporal grounding.}
\method{} improves every grounding metric over the base model at 4B and 8B and gives the best result in 26 of their 28 metrics in Table~\ref{tab:grounding}.
The margin over Video-Zero widens at 8B: on Charades-STA, Video-Zero gains 0.64 mIoU over the base model and \method{} gains 9.40.
Video-Zero derives the temporal evidence it rewards from the model's own predictions, whereas \method{} scores evidence lines against intervals computed by execution, which stay exact at any model scale (Eq.~\ref{eq:evidence_scores}).
Training never sees a real-world interval annotation, so these gains reflect localization transferred from rendered to real videos.

\noindent\textbf{Long-video understanding and video reasoning.}
The gains extend beyond grounding: \method{} improves all eight benchmarks of Table~\ref{tab:qa} on every\ifgemma{} Qwen3-VL\fi{} backbone and gives the best result in 13 of the 16 metrics at 4B and 8B.
Because the training videos contain only simple rendered entities, these results indicate that the temporal skills learned from execution carry over to real-world content.

\label{sec:cross-architecture}

\ifgemma
\noindent\textbf{Other backbones.}
To test whether the recipe depends on the dense Qwen3-VL backbones, we also train Qwen3-VL-30B-A3B-Instruct, a sparse mixture-of-experts model with 3B active parameters, and Gemma~4 E4B~\citep{gemmateam2026gemma4}, from another model family.
In the last two blocks of Tables~\ref{tab:grounding} and~\ref{tab:qa}, \method{} raises 21 of the 22 metrics at 30B-A3B and 21 on Gemma, so its gains rely on neither a dense architecture nor one model family.
\else
\noindent\textbf{Mixture-of-experts backbone.}
To test whether the recipe depends on the dense Qwen3-VL backbones, we apply it to Qwen3-VL-30B-A3B-Instruct, a sparse mixture-of-experts model with 3B active parameters.
In the last block of Tables~\ref{tab:grounding} and~\ref{tab:qa}, \method{} raises 21 of the 22 metrics, by up to 10.78 mIoU on ANet-RTL, so its gains do not rely on a dense architecture.
\fi

\subsection{Ablations}
\label{sec:ablations}

All ablations use Qwen3-VL-4B and remove one component under the same setup (Appendix~\ref{app:control_protocols}); the full method is best in every column.

\noindent\textbf{Questioner.}
The Questioner supplies questions and targets them.
\emph{W/o Questioner} trains the Solver only on the canonical questions that the programs compile, and Overall falls from 50.81 to 47.71, the largest Overall drop.
Proposals ask new questions about the verified criteria of the same instances, whose ground truth stays fixed, so they widen the training distribution without adding label noise.
Targeting comes from the two terms of $\rQ$, of which learnability matters more: \emph{w/o learnability reward} lowers Temporal Grounding by 1.75 points, more than three times the drop of \emph{w/o novelty reward}.
Questions at intermediate pass rates yield Solver groups whose rollouts disagree, so the answer reward separates them, whereas novelty only spreads proposals over task signatures.

\begin{wraptable}{r}{0.54\textwidth}
\vspace{-4pt}
\captionsetup{font=footnotesize,justification=raggedright,singlelinecheck=false}
\caption{\textbf{Ablations on Qwen3-VL-4B.}
Temporal Grounding and Video QA \& Reasoning are the Avg.\ columns of Tables~\ref{tab:grounding} and~\ref{tab:qa}, and Overall averages all twelve benchmarks; marks as in Table~\ref{tab:grounding}.}
\label{tab:ablation}
\centering
\scriptsize
\setlength{\tabcolsep}{2.4pt}
\renewcommand{\arraystretch}{1.0}
\begin{tabularx}{\linewidth}{X ccc}
\toprule
\textbf{Variant} & \shortstack{\textbf{Temporal}\\\textbf{Grounding}} & \shortstack{\textbf{Video QA}\vphantom{g}\\\textbf{\& Reasoning}} & \textbf{Overall} \\
\midrule
\multicolumn{4}{l}{\textit{Questioner}} \\
\quad w/o novelty reward & 36.53 & 56.88 & 50.10 \\
\quad w/o learnability reward & 35.25 & 57.27 & 49.93 \\
\quad w/o Questioner & 31.86 & 55.63 & 47.71 \\
\multicolumn{4}{l}{\textit{Solver}} \\
\quad w/o evidence score & 35.34 & \second{57.33} & 50.00 \\
\quad w/o grounded score & \second{36.63} & 57.11 & \second{50.28} \\
\quad Answer reward only &  31.30 & 56.28 & 47.96 \\
\multicolumn{4}{l}{\textit{Environment evolution}} \\
\quad w/o rule selection & 35.23 & 56.58 & 49.46 \\
\midrule
\ourrow \method{} (full) & \best{37.00} & \best{57.71} & \best{50.81} \\
\bottomrule
\end{tabularx}
\vspace{-1pt}
\end{wraptable}

\noindent\textbf{Solver.}
Evidence terms teach the Solver to locate events.
With \emph{answer reward only}, $\rS=r_{\mathrm{ans}}$ and evidence lines earn no reward, so Temporal Grounding falls by 5.70 points, the largest drop in its column.
Of the two evidence terms, the statement check carries most of this signal.
\emph{W/o evidence score} leaves a correct statement credited only through the interval overlap of $r_{\mathrm{grd}}$ and lowers Temporal Grounding by 1.66 points, over four times the drop of \emph{w/o grounded score}, whose interval credit refines the statement check.
The benefit reaches beyond grounding, since Video QA \& Reasoning also drops the most among the Solver variants when only the answer is rewarded.

\noindent\textbf{Environment evolution.}
Evolution changes which instances are drawn and what happens in them, and \emph{w/o rule selection} removes only the second: it keeps the initial rules, while the curriculum still reallocates instances and the playback schedule still lengthens videos.
Rule changes add event dynamics that reallocation over fixed rules does not supply: without them, Temporal Grounding drops by 1.77 points\ifgemma.\else, and \S\ref{sec:analysis} tests the resulting data apart from the training loop.\fi

%% file: sections/06-analysis.tex
\begin{figure}[t]
\centering
\includegraphics[width=\linewidth]{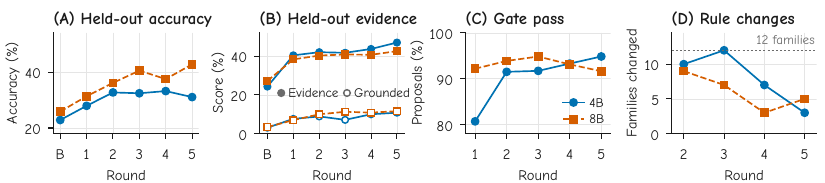}
\caption{\textbf{Training dynamics.}
(A)~Held-out accuracy.
(B)~Evidence (filled) and grounded (hollow) scores; B is the base model.
(C)~Proposals passing the gate.
(D)~Families whose rules change for each round.}
\label{fig:analysis}
\vspace{-2pt}
\end{figure}

\subsection{Analysis}
\label{sec:analysis}

\noindent\textbf{Training dynamics.}
Trained on execution-grounded rewards, both roles improve over the rounds of the 4B and 8B runs (Appendix~\ref{app:analysis_details}).
On 128 held-out instances (Appendix~\ref{app:holdout}), Solver accuracy rises from 25.9\% to 42.8\% at 8B, and the evidence score nearly doubles at 4B, from 24.3\% to 47.0\% (Figure~\ref{fig:analysis}A--B).
\ifgemma\else The grounded score, which credits an interval only on a line that states the correct event, more than triples at both backbones, so later rounds teach the Solver where the cited events occur as well as what they are.
\fi
The share of Questioner proposals that pass the gate rises from 80.7\% to 94.9\% at 4B and stays above 91\% at 8B (Figure~\ref{fig:analysis}C), so the policy learns the gate's validity checks.
Rule selection is most active early: the policy changes rules in 56 of its 96 family decisions, most of them for rounds 2 and 3 (Figure~\ref{fig:analysis}D), and every change stays verifiable, since all menu alternatives compile into instances with recomputed criteria\ifgemma\else, as in Figure~\ref{fig:challenges}b\fi{} (Appendix~\ref{app:rule_decisions}).

\pagebreak
\begin{wrapfigure}{r}{0.46\textwidth}
\vspace{-10pt}
\centering
\captionsetup{font=footnotesize,justification=raggedright,singlelinecheck=false}
\includegraphics[width=0.80\linewidth]{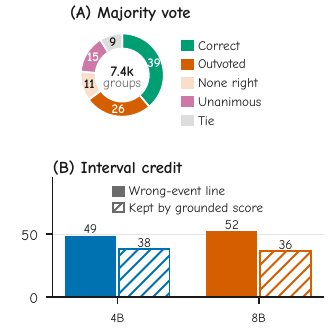}
\caption{\textbf{Execution labels.}
(A)~Majority vote versus execution labels on 7,420 Solver groups (\%).
(B)~Rollouts credited only on wrong-event lines, and credit kept by $r_{\mathrm{grd}}$ (\%).}
\label{fig:analysis-labels}
\vspace{10pt}
\captionsetup{font=footnotesize,justification=raggedright,singlelinecheck=false}
\captionof{table}{\textbf{Supervised fine-tuning on bank data.}
\ifgemma The 4B base model fine-tuned for one epoch on the banks of runs without or with rule selection.\else The 4B base model fine-tuned for one epoch on the banks of the run without rule selection (initial rules) or the full run (evolved rules).\fi}
\label{tab:env-source}
\centering
\scriptsize
\setlength{\tabcolsep}{2.4pt}
\renewcommand{\arraystretch}{1.06}
\begin{tabularx}{\linewidth}{X ccc}
\toprule
\textbf{Training data} & \shortstack{\textbf{Temporal}\\\textbf{Grounding}} & \shortstack{\textbf{Video QA}\vphantom{g}\\\textbf{\& Reasoning}} & \textbf{Overall} \\
\midrule
Base model & 25.08 & 54.68 & 44.81 \\
Initial rules & \second{30.73} & \second{56.71} & \second{48.05} \\
\ourrow Evolved rules & \best{31.78} & \best{57.09} & \best{48.65} \\
\bottomrule
\end{tabularx}
\vspace{1pt}

\end{wrapfigure}

\noindent\textbf{Execution labels versus majority voting.}
These gains rest on execution labels, which can replace consensus only if they hold on the frames that the policy sees, as the checks confirm: recomputation discards fewer than 1\% of the compiled criteria (Appendix~\ref{app:check_stats}), and every Solver group is scored against a criterion that passed both checks, so its label matches the sampled frames (Appendix~\ref{app:consensus}).
Majority voting, a labeling rule of prior video self-play, would assign a wrong label to 52.5\% of 7,420 Solver groups (Figure~\ref{fig:analysis-labels}A).
In 26.3\% of all groups a rollout chooses the correct option but loses the vote, as in Figure~\ref{fig:challenges}a, and in 14.8\% all eight rollouts agree on a wrong option, which agreement filtering would keep.
The wrong-label share is 56.8\% at 4B and 48.2\% at 8B, so consensus labels would mislabel about half of the training questions at both scales.
Evidence credit shows the same contrast: about half of the rollouts that the content-blind overlap $\max_k\mathrm{IoU}(I_k,\gtint_c)$ would credit earn it only on lines that state another event, and $r_{\mathrm{grd}}$ withholds that credit (Figure~\ref{fig:analysis-labels}B).

\noindent\textbf{Training value of evolved environments.}
Whether the event dynamics that rule selection adds also yield better training data can be tested apart from the training loop, by fine-tuning the 4B base model on the banks of each run (Table~\ref{tab:env-source}; Appendix~\ref{app:task-value}).
Both sources improve over the base model in all columns, so execution-labeled data transfer to real videos even outside the self-play loop.
Data from the evolved rules score higher in every column, so the data produced under rule selection transfer better.
\ifgemma\else Because the two runs share the curriculum and training budget, this gap reflects the data that rule selection produces, and removing rule selection also lowers Temporal Grounding by 1.77 points in Table~\ref{tab:ablation}.
\fi

%% file: sections/07-conclusion.tex
\section{Conclusion}
\label{sec:conclusion}

We presented \method{}, a self-improvement framework that grounds video self-play in an executable program library, whose programs render synthetic videos together with ground-truth options and intervals, so supervision stays verifiable without external video corpora or majority voting.
A single policy acts as both Questioner and Solver and is trained on execution-grounded rewards, while a curriculum allocates tasks by pass rate and the policy selects rule changes.
Trained only on simple programs, without external video data or supervised fine-tuning, \method{} outperforms base models and recent self-play baselines on average across twelve real-world benchmarks at different model scales, especially in temporal grounding.
\label{main:end}

%% file: sections/08-statements.tex
\section*{Ethics Statement}
All training videos are rendered from synthetic programs, so training uses no human subjects, personal data, or scraped video.
Evaluation uses twelve public benchmarks under their released protocols.
Appendix~\ref{app:limitations} discusses limitations and broader impact.

\section*{Reproducibility Statement}
Algorithm~\ref{alg:tempo} in Appendix~\ref{app:algorithm} gives the self-play loop, and Table~\ref{tab:hyperparameters} lists the training configuration shared by the 4B, 8B, \ifgemma 30B-A3B, and Gemma\else and 30B-A3B\fi{} models.
Appendices~\ref{app:cost} and~\ref{app:panels} describe compute and the evaluation protocol.
Appendix~\ref{app:environments} specifies the twelve environment families, frame sampling, and the input and output protocols of both roles, and Appendix~\ref{app:rewards} defines the rewards, the question bank, and the optimization.
Appendix~\ref{app:controls} defines the ablation variants, and Appendix~\ref{app:prompts} lists the prompts together with a Solver response.
Code will be made publicly available.

\section*{AI Use Statement}
We used generative AI tools to draft and edit the paper text, to format references, tables, and figures, and to create and edit software code.
The authors reviewed all AI-assisted text and code and take responsibility for the full content of this work, including text, claims, and artifacts produced with the aid of generative AI.

%% file: sections/A-appendix.tex
\section{Training Configuration and Evaluation}
\label{app:protocol}

\subsection{Self-Play Algorithm}
\label{app:algorithm}

Algorithm~\ref{alg:tempo} outlines the training loop of the shared vision-language policy.
$\textsc{Batch}$ draws the next batch from the allocation that the curriculum of \S\ref{sec:evolution} fixes for the round (Eq.~\ref{eq:weight}) and renders it ahead of the step that uses it.
The bank stores every measured canonical question and at most one gated proposal per instance, preselected by task-class and task rarity.
At capacity, eviction first reduces overrepresented task classes and families and then removes the oldest measurements (Appendix~\ref{app:bank}).
A single AdamW optimizer state persists across both roles and across rounds.

\input{sections/algorithm}

\subsection{Training Configuration}
\label{app:hyperparameters}

Table~\ref{tab:hyperparameters} lists the settings shared by the 4B, 8B, \ifgemma 30B-A3B, and Gemma\else and 30B-A3B\fi{} configurations.
The Questioner and Solver contribute separate reward groups to one combined update of the LoRA adapters on the language model, and rule selection after each round runs at inference only.
The instruction-tuned checkpoint, obtained by disabling the adapters, serves as the reference policy for KL regularization.

\begin{table}[t]
\caption{\textbf{Training configuration.}
Settings shared by the 4B, 8B, \ifgemma 30B-A3B, and Gemma\else and 30B-A3B\fi{} configurations.}
\label{tab:hyperparameters}
\centering\small
\renewcommand{\arraystretch}{1.12}
\begin{tabularx}{\linewidth}{@{}l >{\raggedright\arraybackslash}X@{}}
\toprule
\textbf{Component} & \textbf{Setting} \\
\midrule
Backbone & Qwen3-VL-4B/8B/30B-A3B-Instruct\ifgemma{} and Gemma-4-E4B-it\fi; LoRA with rank 64 and $\alpha=128$ on the q/k/v/o projections of the language model; the vision tower and the A3B experts carry no adapter \\
Environments & 12 environment families, 3 difficulty levels; 5 rounds of 20 steps \\
Batch & 32 instances per step; playback scale $\lambda_r=1.0, 1.25, 1.5, 1.75, 2.0$ for rounds $r=1,\ldots,5$ \\
Questioner & 4 proposals per instance; one to three located evidence lines \\
Solver & $M=8$ rollouts per question; one to eight evidence lines; criteria are those of the events the target rests on, or the target itself when it rests on none; at most 72 groups per step (32 canonical, up to 32 preselected proposals, 8 replayed) \\
Question bank & Capacity 640, shared across rounds; up to 8 replayed questions per step; at most one preselected proposal per instance; one prior example in the Questioner prompt \\
Option count & Cycles across (6, 6, 6, 5, 4) \\
Curriculum & Coverage share $\kappa=0.30$; pass-rate decay 0.90 per step; level thresholds $\rho_{\mathrm{adv}}=0.80$ and $\rho_{\mathrm{ret}}=0.20$ on the decayed pass rate, applied with at least 3 questions; class weights start equal and change by at most a factor of 2 per round \\
Rule menu & At most 3 alternative configurations besides keep; one selection per family per round \\
Optimizer & AdamW, learning rate $5\times10^{-5}$, weight decay $10^{-2}$ \\
Schedule & Constant learning rate without warmup; single PPO epoch per step; one combined update over both roles, 100 updates per run \\
Regularization & Low-variance KL with coefficient $\beta_{\mathrm{KL}}=10^{-3}$ against the base model with adapters disabled; clipping threshold 0.2; gradient norm cap 0.5; advantages mean-centered without standard-deviation normalization \\
Precision & fp32 master weights; bf16 compute and rollout \\
Sampling & Temperature 1.0; top-$p$ 0.99; unrestricted top-$k$ \\
Token limits & 16,384 prompt tokens; 2,048 response tokens; context ceiling 24,576 \\
Visual input & Uniform sampling at 2 fps, 4 to 64 frames, so clips longer than 32\,s fall below 2 fps; 3,136 to 100,352 pixels per frame \\
Checkpointing & Saved after every round (every 20 steps) \\
Holdout & 128 synthetic instances, evaluated before training and after every round; used for neither training nor curriculum \\
Hardware & 8 GPUs per run \\
\bottomrule
\end{tabularx}
\end{table}

The scheduled budget comprises 3,200 instances and at most 12,800 Questioner proposals.
Every canonical question, gated proposal, and replayed question is answered eight times ($M=8$) under the policy snapshot that generated the proposals.
These responses give every gated proposal its pass rate for the Questioner reward, and the responses to canonical questions, preselected proposals, and replayed questions also train the Solver.
Groups whose rewards do not differ supply pass rates but no gradient, and both roles are scored before the shared parameters change.

\subsection{Compute Resources}
\label{app:cost}

Each training run uses eight GPUs.
A single persistent inference engine serves rollouts for both roles, while video rendering runs asynchronously in a CPU worker pool scheduled one step ahead of the policy update.

\subsection{Evaluation Protocol}
\label{app:panels}

\noindent\textbf{Benchmark suites.}
Video reasoning is evaluated on LongVideoReason~\citep{chen2025longvideoreason}, MMVU~\citep{zhao2025mmvu}, VideoMathQA~\citep{rasheed2025videomathqa}, and VideoMMMU~\citep{hu2026videommmu}, scored by question-answering accuracy (\%).
Temporal grounding is evaluated on ANet-RTL~\citep{huang2024lita}, ActivityNet Captions~\citep{krishna2017activitynet}, Charades-STA~\citep{gao2017tall}, and NExT-GQA~\citep{xiao2024nextgqa}.
The first three report mean Intersection over Union (mIoU) between predicted and ground-truth intervals together with recall at IoU thresholds 0.3, 0.5, and 0.7, and NExT-GQA reports grounded accuracy at IoU 0.3 and 0.5, which requires both a correct answer and temporal overlap.
Long-video understanding is evaluated on LongVideoBench~\citep{wu2024longvideobench}, MLVU~\citep{zhou2025mlvu}, Video-MME-Long~\citep{fu2025videomme}, and LSDBench~\citep{qu2025lsdbench}, which probe visual context over extended durations and are scored by question-answering accuracy (\%).

\noindent\textbf{Final evaluation.}
Final numbers use the checkpoint from the fifth training round on the complete test splits of all twelve benchmarks.
All models, including the baselines and ablation variants, are evaluated under the same protocol, which follows~\citet{zhang2026videozero} and the official metrics of each benchmark.

\section{Executable Environments and Interfaces}
\label{app:environments}

\subsection{Environment Coverage and Design}
\label{app:env_coverage}

\begin{figure}[t]
\centering
\includegraphics[width=\linewidth]{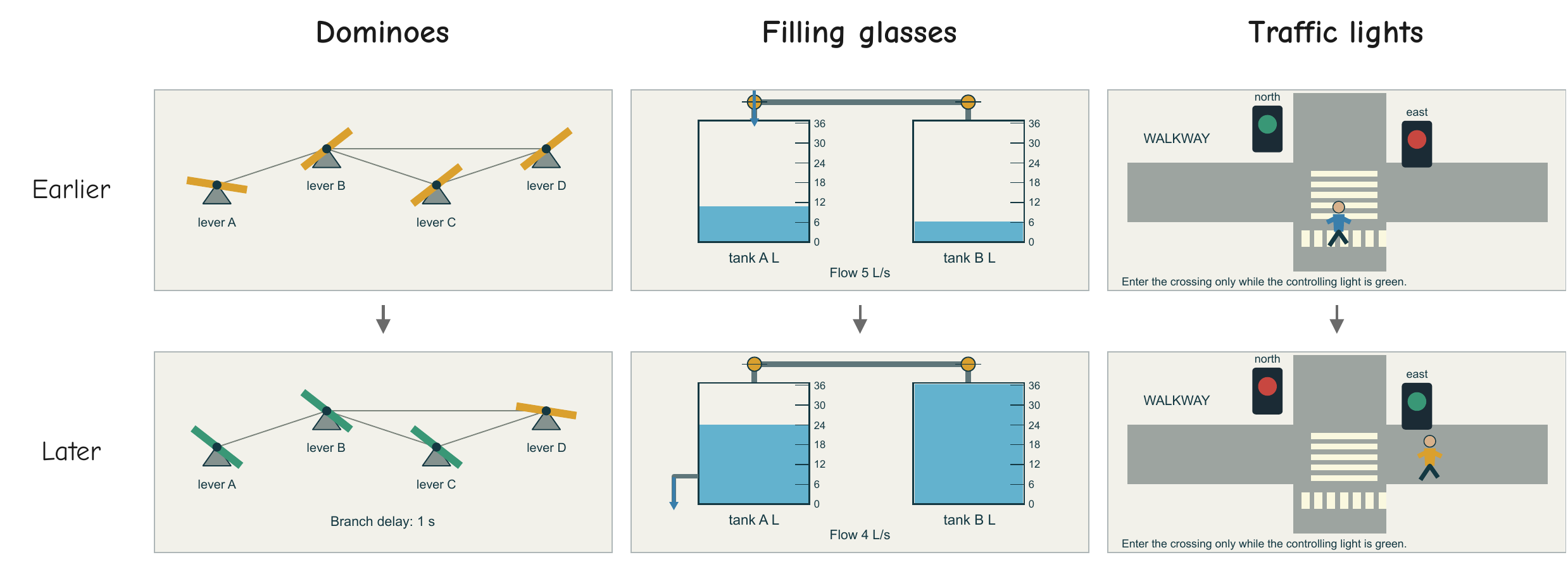}
\caption{\textbf{Executable environments.}
Rendered frame pairs from Dominoes, Filling glasses, and Traffic lights; the labels inside the frames are part of the rendered video.}
\label{fig:program}
\end{figure}

Table~\ref{tab:tasks} lists the tasks by class, and Table~\ref{tab:families} summarizes the program library.
\begin{table}[t]
\caption{\textbf{Tasks by class.}
Tasks marked $^\dagger$ are interval questions, answered with a free-form interval; all other tasks are multiple-choice.}
\label{tab:tasks}
\centering\scriptsize
\begin{tabularx}{\linewidth}{@{}l >{\raggedright\arraybackslash}X@{}}
\toprule
\textbf{Class} & \textbf{Tasks}\\
\midrule
Interval & Event grounding$^\dagger$, state interval$^\dagger$, closest interval$^\dagger$, tile motion interval$^\dagger$, interval relation, interval composition\\
Event sequence & Event order, adjacent event, path sequence, pattern extrapolation\\
Quantity comparison & Event count, extremum, temporal extremum, rate trend, speed comparison, closest pair, travel rank\\
Motion & Motion direction, relative motion, motion profile, trajectory shape, distance trend\\
State inference & Forward state, backward state, history query, occluded identity, tile endpoint, tile history, tile waypoint\\
Container update & Set update, multiset update, sequence update\\
Numeric computation & State change, arithmetic composition, expression evaluation\\
\bottomrule
\end{tabularx}
\end{table}
Each environment family implements a deterministic state-transition system, rendering routines, query templates, and analytical verification routines.
Difficulty levels govern the number of interacting entities and the operation complexity, while the playback scale sets the temporal duration.
Levels enter in rotation and thereafter follow the curriculum, and option counts cycle over a fixed sequence.
Figure~\ref{fig:program} shows state transitions from three representative families.

\begin{table}[t]
\caption{\textbf{Twelve environment families.}
Each family supports three difficulty levels; the columns summarize its observable dynamics and the reasoning targets it supports.}
\label{tab:families}
\centering\small
\renewcommand{\arraystretch}{1.12}
\begin{tabularx}{\linewidth}{@{}l >{\raggedright\arraybackslash}X >{\raggedright\arraybackslash}X@{}}
\toprule
\textbf{Family} & \textbf{Observable dynamics} & \textbf{Reasoning targets} \\
\midrule
Billiards & Kinematic motion, elastic contact, and velocity shifts & Trajectory direction, event order, state transitions \\
Block stacking & Sequential placement and physical support & Object counts, stacking order, spatial relations \\
Checkout queue & Arrival, service progression, and departure & Queue duration, entity counts, interval relations \\
Dominoes & Cascading propagation along contact dependencies & Chain continuation, trigger ordering, collapse speed \\
Ferris wheel & Coupled rotating structures and cyclic cabins & Relative angular rates, phase shifts, event timing \\
Filling glasses & Liquid level shifts and volumetric transfers & Continuous quantities, extrema, volume change intervals \\
Maze run & Discrete path progression and visited markers & Route reconstruction, intersection counts, milestone order \\
Parking lot & Vehicle ingress, stall occupancy, and egress & Occupancy states, dwell durations, interval intersections \\
Shell game & Continuous occlusion, permutations, and transfers & Object tracking, identity recovery, hidden states \\
Tag spread & Contact-induced state propagation across agents & Infection ordering, transmission delays, final states \\
Traffic lights & Periodic signal phases and permitted trajectories & Phase duration counts, rule compliance, interval localization \\
Tunnel trains & Motion across visible and occluded segments & Transit speed, occluded intervals, exit order \\
\bottomrule
\end{tabularx}
\end{table}

\subsection{Video Frame Sampling and Temporal Coordinates}
\label{app:sampling_coords}

Frames are sampled uniformly over the video at a fixed target rate up to a fixed maximum count, always including the initial and final states.
A criterion's target must be visible in at least three sampled frames.
Every ground-truth value is recomputed from the sampled frames, and a criterion is discarded when its value cannot be recomputed from those frames or disagrees with the simulation.
The static-question screen records three outcomes: changed under an evaluable transformation, unchanged under both evaluable transformations, or undetermined.
The second outcome is excluded, and undetermined criteria remain subject to the normal visibility and validity checks.
A transformation cannot be evaluated when reversal creates tied events or leaves the query's supported domain.
Criteria of tasks that declare every operation they depend on, such as set update and expression evaluation, are recorded as undetermined without running the transformations, because each of those operations is already recomputed from the sampled frames.

All ground-truth temporal labels use the effective time coordinate exposed to the model.
For $F$ selected frames spanning the rendered video duration $d$, which playback scale $\lambda_r$ stretches, the observation manifest defines
\begin{sequation}
f_{\mathrm{label}} = \frac{F-1}{d}, \qquad t_i = \frac{i}{f_{\mathrm{label}}}, \quad i = 0, \ldots, F-1.
\label{eq:label_clock}
\end{sequation}
The input processor consumes this effective frame rate and the selected frames directly, without a second temporal downsampling, so verification intervals map to the same discrete timestamps that the model sees.

\subsection{Verification Checks on the Training Environments}
\label{app:check_stats}

\begin{figure}[htbp]
\centering
\includegraphics[width=0.62\linewidth]{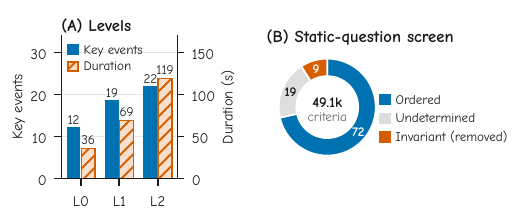}
\caption{\textbf{Training environments.} (A)~Key events and duration by level. (B)~Static-question screen classes in \%; only invariant criteria are removed.}
\label{fig:env-checks}
\end{figure}

To measure how often the two checks of \S\ref{sec:environments} apply, we read the check records of the 5,074 distinct instances that the three \ifgemma Qwen3-VL training runs\else training runs, one per backbone,\fi{} draw from the twelve families, 2,599 at level 0, 1,179 at level 1, and 1,296 at level 2.
A level-2 video holds 22 key events over 119 seconds on average, against 12 over 36 seconds at level 0 (Figure~\ref{fig:env-checks}A).
Frames are selected uniformly over each video, independently of where events occur, at no more than 2\,fps and 64 frames per video.
A level-0 video thus has 58.1 frames on average at an effective rate of 1.69\,fps, and a level-2 video 64.0 frames at 0.62\,fps (Eq.~\ref{eq:label_clock}).
Table~\ref{tab:check_stats} shows that the share of recorded events that start and end between two sampled frames grows from 2.82\% at level 0 to 28.56\% at level 2.
A label read from the simulation clock would therefore often describe a state that no sampled frame shows.

Two parts of the environment prevent this.
First, the key events that questions refer to are limited to events that at least one sampled frame shows and that lie at least two sampled frames apart, and an instance must keep at least 3, 5, or 7 such events at levels 0, 1, and 2.
Every one of the 82,392 key events in these instances therefore appears on a sampled frame.
Second, the recomputation discards 403 of the 49,525 compiled criteria: in 208 the recomputed value disagrees with the simulation, and 195 cannot be recomputed from the frames.
Disagreements concentrate in Event count and Extremum, with 110 and 68 criteria, and the most frequent case counts three events on the simulation clock, of which the sampled frames show two.
The discarded share grows from 0.19\% at level 0 to 2.13\% at level 2, as the share of events between frames does.

The static-question screen removes only 9.01\% of the remaining 49,122 criteria, those whose answer stays the same under both reversal and permutation (Figure~\ref{fig:env-checks}B).
Nearly all criteria of Event grounding, Event order, and State interval pass the screen, with at most 0.13\% removed.
The removed criteria come mainly from Extremum and Event count, where 35.69\% and 34.04\% of each task's criteria keep the same answer under both transformations, since many counts and comparisons do not depend on event order.

\begin{table}[htbp]
\caption{\textbf{Verification checks on the training environments.}
Instances per level, mean sampled frames and effective frame rate, share in \% of recorded events that fall between two sampled frames, criteria compiled and discarded by the recomputation, and shares in \% of the static-question screen classes among the criteria that pass the recomputation.
Invariant criteria are removed, and ordered and undetermined criteria are kept.}
\label{tab:check_stats}
\centering\footnotesize
\setlength{\tabcolsep}{3.0pt}
\renewcommand{\arraystretch}{1.08}
\begin{tabular}{@{}l r cc c r rr ccc@{}}
\toprule
& & \multicolumn{2}{c}{\textbf{Sampled frames}} & \textbf{Events} & \multicolumn{3}{c}{\textbf{Recomputation}} & \multicolumn{3}{c}{\textbf{Static-question screen}} \\
\cmidrule(lr){3-4}\cmidrule(lr){6-8}\cmidrule(l){9-11}
\textbf{Level} & \textbf{Instances} & Count & fps & \textbf{between} & Criteria & Disagree & Unverified & Ordered & Invariant & Undet. \\
\midrule
0 & 2,599 & 58.1 & 1.69 & 2.82 & 25,338 & 23 & 25 & 67.93 & 10.76 & 21.31 \\
1 & 1,179 & 63.8 & 1.00 & 10.02 & 11,623 & 44 & 44 & 74.33 & 7.29 & 18.38 \\
2 & 1,296 & 64.0 & 0.62 & 28.56 & 12,564 & 141 & 126 & 76.21 & 7.02 & 16.77 \\
\midrule
Overall & 5,074 & 60.9 & 1.26 & 16.60 & 49,525 & 208 & 195 & 71.51 & 9.01 & 19.49 \\
\bottomrule
\end{tabular}
\end{table}

\subsection{Role Inputs and Output Protocols}
\label{app:protocols}

The Questioner receives the rendered frames, the scene description and active rules from the execution metadata, the askable criteria with their ground-truth options and distractor domains, one question of the same task from the bank, and the Solver's pass rate per criterion.
No timestamp of an event that a criterion rests on is shown, so the Questioner reads the intervals of its located lines from the frames, and novelty reward $r_{\mathrm{nov}}$ checks them (Appendix~\ref{app:signals}).
Each criterion specifies a target entity and the relation evaluated on it, a question stem, the ground-truth option, the ground-truth interval, and a distractor domain with a resolution floor.
The Solver receives only the frames, the question text, and, for a multiple-choice question, the options.

When proposing, the Questioner names one verified criterion, writes the stem and the options, and cites the events the criterion depends on as located evidence lines before its answer payload.
The gate accepts a proposal only if exactly one option equals the ground-truth option $y^\star$ and all distractors lie in the criterion's domain, separated at the declared temporal or numerical resolution.
The stem must also match the task specification, name the criterion's subjects, keep its qualifiers, state no ground-truth value, and end in a single question mark.
The environment runtime then determines the final option permutation and the answer letter.
Criteria without enough distinct distractors are excluded from the askable set, and the Questioner's located lines are never shown to the Solver.

The Solver is prompted to write located evidence lines, then derivation lines without intervals, and finally a single option character inside \texttt{<answer>}, or for an interval question the start and end seconds of the interval as JSON.
Each evidence line starts with a list of \texttt{[start, end]} pairs in decimal seconds, followed by one observed event or state; a recurring identical event or state places all of its intervals in that one list.
Static in-context demonstrations define these output structures (Appendix~\ref{app:prompts}).

\subsection{Rule Selection and Execution Determinism}
\label{app:rule_selection}

After each round except the last, the environment runtime composes for every family a menu of at most three alternative rule configurations besides the current one.
Each entry moves one rule to a value the family has not run, and from the third round on, entries that move a pair of rules lead the menu.
The policy receives the menu with a preview of each entry, the rule meanings, the current executable graph, the editable schema, the current class weights, prior changes, and measurements under the current configuration and playback scale.
Canonical and Questioner measurements remain separate, and recorded responses and evidence scores are provided as feedback.
The output selects a menu entry and sets the class weights of the family, which enter $w_{f,a}$; Shell game additionally supports operation-list, dependency, and operation-parameter edits of its operation graph.
Each weight changes by at most a factor of two per round, and only for classes with at least three current-configuration questions.
The interface permits neither arbitrary program rewriting nor independent changes to ground truth.
The executor checks field validity, parameter domains, references, finite expansion, and acyclicity, and a joint preview executes the proposed configuration and graph together before rendering and recompiling the resulting supervision.
Invalid or failed outputs keep the previous program.
Task levels move by the pass-rate thresholds of \S\ref{sec:evolution} and class weights by the policy's bounded selection.
A task that produced no instance in a round although at least three were requested steps down one level.
Interval tasks instead alternate between two tiers, level 0 at playback scale 1.0 and level 2 at playback scale 2.0, and the state-tracking tasks, namely set, multiset, and sequence update, occluded identity, history query, and expression evaluation, take levels 0 and 1 in the first 20 steps, levels 1 and 2 in the next 20 steps, and level 2 afterward.
Task weight $w_{f,a}$ is the class weight of $a$ times a fixed tilt of 2.6 for interval tasks, 4.0 for state-tracking tasks, 0 for tile endpoint, tile waypoint, tile history, and forward state, and 1 otherwise, and every task keeps at least 4 percent of its family's instances.

Selected configurations and program graphs determine future traces, renderings, criteria, and answers.
Instance records preserve the original generation context: replay resolves the saved instance identifier through its frame manifest and keeps the question's original scoring context, so later environment edits do not alter earlier questions.
The curriculum constants, namely the pass-rate decay, the coverage share $\kappa$, the level thresholds and their minimum question count, and the weight bound, are fixed across runs (Table~\ref{tab:hyperparameters}).

\begin{figure}[htbp]
\centering
\small
\textit{Before: specular side cushions}\\[2pt]
\includegraphics[width=\linewidth]{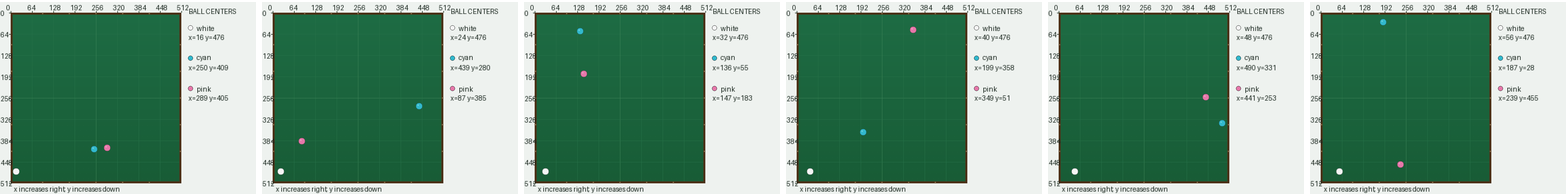}\\[4pt]
\textit{After: retroreflective side cushions}\\[2pt]
\includegraphics[width=\linewidth]{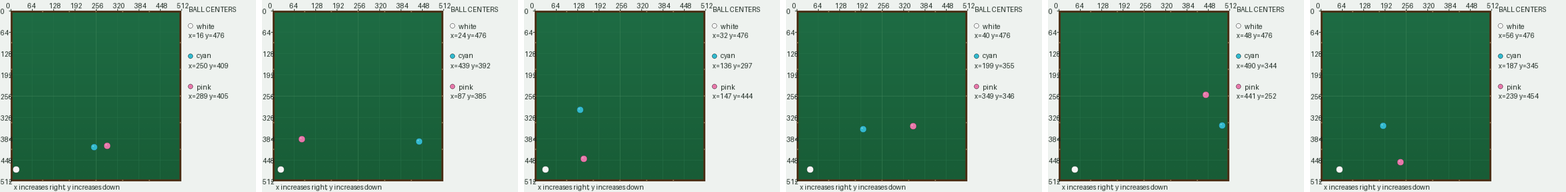}
\caption{\textbf{A rule change in Billiards.}
Frames at 0.0, 4.0, 7.9, 11.9, 15.9, and 19.8\,s of the same instance before and after the rule change; the overlays are part of the rendered frames.}
\label{fig:rule_case}
\end{figure}

\noindent\textbf{Example of a rule change.}
Figure~\ref{fig:rule_case} shows one rule change in Billiards with the seed and level held fixed.
Changing the side cushion from specular to retroreflective reflection sends the pink ball back along its incoming path.
The canonical question, ``Which table quadrants does the pink ball enter on its first 12 quadrant changes, in order?'', stays the same, but re-execution changes its ground-truth option from ``lower-left, upper-left, upper-right, lower-right'' repeated three times to ``lower-left, lower-right, upper-right, lower-right'' repeated three times, together with the 12 ground-truth intervals.
The sampled frames coincide up to 4.0\,s and differ from 7.9\,s on, so the new rule yields new event dynamics whose labels remain exact.

\section{Rewards, Question Bank, and Optimization}
\label{app:rewards}

\subsection{Reward Terms}
\label{app:signals}

This section details the reward terms of Eq.~\ref{eq:rewards}.

\noindent\textbf{Learnability.}
Let $s$ of the $M$ Solver rollouts on question $q$ be correct, where an interval answer is correct when its IoU with $\gtint_{c_q}$ is at least $\epsilon=0.5$.
With pass rate $p=s/M$ as the estimated probability that one rollout is correct, the Bernoulli variance $p(1-p)$ peaks at $p=1/2$ and vanishes when all rollouts are correct or all are wrong.
Laplace smoothing replaces $p$ by $\tilde p=(s+1)/(M+2)$, and the learnability reward
\begin{sequation}
r_{\mathrm{learn}}(q)=4\,\tilde p\,(1-\tilde p)=\frac{4(s+1)(M-s+1)}{(M+2)^2}
\label{eq:learn_deriv}
\end{sequation}
equals one at $s=M/2$ and stays positive at $s\in\{0,M\}$.
With $M=8$, it takes the values 0.36, 0.64, 0.84, 0.96, and 1 for $s=0,1,2,3,4$ and is symmetric around $s=4$.

\noindent\textbf{Evidence matching.}
The Solver's evidence lines are parsed with the grammar of Appendix~\ref{app:protocols}.
Each line is assigned to at most one criterion of $\mathcal{C}_q$ and each criterion to at most one line, so that the assignment is one-to-one; $k(c)$ denotes the line assigned to criterion $c$.
Among such assignments, the one with the most matched criteria, and then the highest grounded score defined below, is kept.
The indicator $\delta_c$ equals one when the assigned line states the event of $c$ and zero otherwise; a criterion without an assigned line has $\delta_c=0$.
A line states the event of $c$ when it asserts the recorded value of $c$ without negation, in a clause that names a subject of $c$ and a term of its relation.
An ordered value must list its items in the recorded order, and a duration must be stated as an interval.
With $m=\sum_{c\in\mathcal{C}_q}\delta_c$ matched criteria, precision $P=m/L$ and recall $R=m/|\mathcal{C}_q|$ give
\begin{sequation}
r_{\mathrm{evid}}=\frac{2PR}{P+R}=\frac{2m}{L+|\mathcal{C}_q|},
\end{sequation}
which is zero when no criterion is matched.
The grounded score averages $\delta_c\,\mathrm{IoU}(I_{k(c)},\gtint_c)$ over $\mathcal{C}_q$, where the IoU of two finite unions of intervals is the length of their intersection divided by the length of their union.
In this score, $\gtint_c$ is widened by a factor $\zeta=1.4$ about the center of each of its intervals and clipped to the video.
A line that states a verified fact of the instance outside $\mathcal{C}_q$ is left out of $L$, so a correct observation beyond the scored criteria costs no precision.
When the criterion of $q$ declares alternative sets of supporting criteria, each set is scored as $\mathcal{C}_q$, and the set with the highest $r_{\mathrm{evid}}+r_{\mathrm{grd}}$ is kept.

\noindent\textbf{Novelty.}
The novelty reward credits the information that a proposal adds to its input, namely the intervals of its evidence lines, since the Questioner sees no timestamp of the events that a criterion rests on.
It applies the grounded score of Eq.~\ref{eq:evidence_scores} to these lines,
\begin{sequation}
r_{\mathrm{nov}}(q)=\mean_{c\in\mathcal{C}^{\mathrm{Q}}_q}\;\delta_c\,\mathrm{IoU}\bigl(I_{k(c)},\gtint_c\bigr),
\label{eq:nov_formula}
\end{sequation}
where $\mathcal{C}^{\mathrm{Q}}_q\subseteq\mathcal{C}_q$ is formed from criterion $c_q$ as for the Solver and leaves out every criterion whose interval is printed in $c_q$, the stem, or the options.
Before scoring, a line whose every interval restates such a printed interval is dropped.
A proposal with $\mathcal{C}^{\mathrm{Q}}_q=\emptyset$ receives $r_{\mathrm{nov}}=0$, so copying a printed interval earns no credit.

\subsection{Question Bank}
\label{app:bank}

Every measured canonical question enters the bank, so each instance contributes at least one stored question.
Among the proposals that pass gate $g(q)$ (\S\ref{sec:selfplay}), the bank preselects at most one per instance, favoring the task class that is rarest in the bank and, within it, the rarest task, and stores it once measured; at most 14 percent of a step's instances contribute an interval question; proposals with $g(q)=0$ are neither answered nor stored.
Each entry keeps the question, its verification context, the rule configuration and program graph of its instance, and the instance identifier.
Replay reloads the rendered frames of the stored instance and answers them with the current policy.
For every (family, task) pair and level, the bank keeps exponentially decayed pass-rate statistics, including counts of solved ($p=1$), learnable ($0<p<1$), and unsolved ($p=0$) questions, which the curriculum of \S\ref{sec:evolution} and the rule-selection prompt read.
The bank holds 640 questions shared across rounds; at capacity, eviction first removes questions of overrepresented task classes and families and then the oldest measurements, which keeps all task classes represented.

\subsection{Group Relative Normalization and Policy Optimization}
\label{app:optimization}
Policy parameters $\theta$ are shared between Questioner $\pi_\theta^{\mathrm{Q}}$ and Solver $\pi_\theta^{\mathrm{S}}$ and updated with Group Relative Policy Optimization (GRPO).
The Solver forms groups $\mathcal{G}_{\mathrm{S}}$ of $M=8$ rollouts per question, and the Questioner forms groups $\mathcal{G}_{\mathrm{Q}}$ of 4 proposals per instance.
Both rewards subtract $\lambda_{\mathrm{dup}}=0.5$ times the share of evidence lines that exactly repeat an earlier line; for the Questioner, this term lies outside the gate.
A response is defective when it reaches the length cap, leaves its answer unclosed, loops, or, for the Solver, writes nothing before its answer.
A defective proposal receives $\rQ=-1$.
A defective rollout receives 0.5 less than the lowest reward among the well-formed rollouts of its group, and its advantage is multiplied by its token count over their mean token count, clipped to $[1,8]$, which restores the per-token weight that the sequence-mean loss takes from a long response.
The advantage of each sample is its reward minus the group mean,
\begin{sequation}
\hat A_i = r_i - \frac{1}{|\mathcal{G}|}\sum_{j\in\mathcal{G}}r_j,
\label{eq:grpo_advantage}
\end{sequation}
without division by the group standard deviation, which keeps the scale of reward differences~\citep{liu2025drgrpo}.
Groups whose rewards are all equal have zero advantage and are omitted from backpropagation.
Both roles use the clipped surrogate objective with a Kullback--Leibler penalty $\beta_{\mathrm{KL}} = 10^{-3}$ against the frozen reference policy $\pi_{\mathrm{ref}}$, the base model with adapters disabled.
Training performs one combined Questioner--Solver update per step with AdamW at learning rate $5\times10^{-5}$.
The shared policy is held fixed throughout proposal sampling, Solver sampling, and reward evaluation, so each Questioner reward measures the current Solver without an intermediate parameter change.
Each role's advantages are centered within its own groups, and the optimizer receives the eligible groups of both roles.

\FloatBarrier
\section{Analysis Details}
\label{app:analysis_details}

The analyses of \S\ref{sec:analysis} use the recorded proposals, rollouts, rule decisions, and held-out evaluations of the 4B and 8B training runs.

\subsection{Majority Voting and Interval Credit by Backbone}
\label{app:consensus}

For every Solver group, majority voting takes the option chosen most often among the eight rollouts, and a tie leaves no majority label.
Table~\ref{tab:consensus} compares this label with the execution label.
Every one of these groups is scored against a criterion that passed the recomputation and the static-question screen of \S\ref{sec:environments}, and its ground-truth value equals the value recomputed on the sampled frames.
Re-executing the programs of all instances reproduces the same traces, frame selections, and discarded criteria, so the execution labels are reproducible.
The majority label is wrong for 56.8\% and 48.2\% of the groups at 4B and 8B and for 50.8\% to 53.5\% at each difficulty level, so it would mislabel about half of the training questions regardless of scale or video length.
Wrong labels on which all eight rollouts agree are most frequent at 4B, where agreement filtering would keep a wrong label for 21.3\% of the groups.
The right half of the table covers all Solver rollouts.
More than 90\% of them receive some content-blind overlap.
In 48.8\% and 52.2\% of these rollouts at 4B and 8B, the overlap comes only from lines whose statement does not match the criterion, so $r_{\mathrm{grd}}$ gives them no interval credit.
The grounded score keeps 36\% to 38\% of the content-blind credit, and the credit it keeps falls only on lines that state the correct event (Eq.~\ref{eq:evidence_scores}).

\begin{table}[htbp]
\caption{\textbf{Majority voting and interval credit.}
Left: Solver groups, with shares in \% of groups whose majority label is wrong, whose majority label is wrong although at least one rollout chooses the correct option, whose eight rollouts agree on a wrong option, and whose vote ties.
Right: Solver rollouts, with the share in \% of rollouts credited by the content-blind overlap whose credit falls only on lines with a non-matching statement, and the share in \% of the content-blind credit that $r_{\mathrm{grd}}$ keeps.}
\label{tab:consensus}
\centering\footnotesize
\setlength{\tabcolsep}{4pt}
\renewcommand{\arraystretch}{1.08}
\begin{tabular}{@{}l r cccc cc@{}}
\toprule
& \multicolumn{5}{c}{\textbf{Majority voting}} & \multicolumn{2}{c}{\textbf{Interval credit}} \\
\cmidrule(lr){2-6}\cmidrule(l){7-8}
\textbf{Backbone} & Groups & Wrong & Outvoted & Unanimous & Tie & Wrong line & Kept \\
\midrule
4B & 3,703 & 56.82 & 23.17 & 21.33 & 7.24 & 48.78 & 38.01 \\
8B & 3,717 & 48.21 & 29.32 & 8.34 & 10.36 & 52.21 & 36.44 \\
\midrule
Pooled & 7,420 & 52.51 & 26.25 & 14.82 & 8.80 & 50.50 & 37.18 \\
\bottomrule
\end{tabular}
\end{table}

\subsection{Held-Out Instances}
\label{app:holdout}

Each run builds its held-out set once from program seeds that no training round reaches.
The 128 instances are balanced over families and levels at playback scales 1.0 and 2.0, and a candidate that fails the generation checks is replaced by another of the same family and level, so model answers never determine membership.
After every round, the policy answers the canonical question of each instance eight times.
An answer counts as correct when it selects the ground-truth option.
Evidence and grounded scores are $r_{\mathrm{evid}}$ and $r_{\mathrm{grd}}$ of Eq.~\ref{eq:evidence_scores}.
Table~\ref{tab:holdout} shows that all three metrics exceed the base model after every round at both backbones.

\begin{table}[htbp]
\caption{\textbf{Held-out instances.}
Accuracy, evidence score, and grounded score in \% on the 128 held-out instances with eight answers each, for the base model (B) and after each of the five rounds; bold marks the best round.}
\label{tab:holdout}
\centering\footnotesize
\setlength{\tabcolsep}{5pt}
\renewcommand{\arraystretch}{1.08}
\begin{tabular}{@{}ll cccccc@{}}
\toprule
\textbf{Backbone} & \textbf{Metric} & B & 1 & 2 & 3 & 4 & 5 \\
\midrule
\multirow{3}{*}{4B} & Accuracy & 22.9 & 27.9 & 32.7 & 32.4 & \best{33.2} & 31.1 \\
& Evidence & 24.3 & 40.4 & 42.1 & 41.7 & 43.7 & \best{47.0} \\
& Grounded & 3.1 & 7.6 & 8.9 & 7.1 & 10.0 & \best{10.8} \\
\midrule
\multirow{3}{*}{8B} & Accuracy & 25.9 & 31.3 & 36.1 & 40.5 & 37.5 & \best{42.8} \\
& Evidence & 27.2 & 38.3 & 40.3 & 40.9 & 40.7 & \best{42.6} \\
& Grounded & 3.4 & 7.0 & 10.0 & 11.3 & 10.8 & \best{11.6} \\
\bottomrule
\end{tabular}
\end{table}

\subsection{Rule Decisions}
\label{app:rule_decisions}

Table~\ref{tab:rule-decisions} lists the 96 family decisions of the two runs.
A decision keeps the rules either because the policy selects the current configuration or because its output fails a check of Appendix~\ref{app:rule_selection}.
Every alternative on the menus, 270 in total, compiles into usable instances in the runtime's preview.

\begin{table}[htbp]
\caption{\textbf{Rule decisions.}
Families whose rules change for each round, decisions that keep the rules by the policy's selection or after a failed check, and menu alternatives with usable previews.
Each backbone makes 12 decisions per round.}
\label{tab:rule-decisions}
\centering\footnotesize
\setlength{\tabcolsep}{4.5pt}
\renewcommand{\arraystretch}{1.08}
\begin{tabular}{@{}l cccc c cc c@{}}
\toprule
& \multicolumn{5}{c}{\textbf{Changed}} & \multicolumn{2}{c}{\textbf{Kept}} & \\
\cmidrule(lr){2-6}\cmidrule(lr){7-8}
\textbf{Backbone} & 2 & 3 & 4 & 5 & Total & Selected & Failed check & \textbf{Usable alternatives} \\
\midrule
4B & 10 & 12 & 7 & 3 & 32 & 12 & 4 & 129 / 129 \\
8B & 9 & 7 & 3 & 5 & 24 & 23 & 1 & 141 / 141 \\
\midrule
Total & 19 & 19 & 10 & 8 & 56 & 35 & 5 & 270 / 270 \\
\bottomrule
\end{tabular}
\end{table}

\FloatBarrier
\section{Controlled Comparisons}
\label{app:controls}

\subsection{Ablation Variants}
\label{app:control_protocols}

Table~\ref{tab:protocol-controls} defines the ablation variants of Table~\ref{tab:ablation}.
Every variant starts from the same initialization and uses the same training budget, curriculum constants, and evaluation protocol as the full method, and it differs from the full method in exactly one component.

\begin{table}[htbp]
\caption{\textbf{Ablation variants.}
Each variant changes one component of the full method.}
\label{tab:protocol-controls}
\centering\small
\renewcommand{\arraystretch}{1.1}
\begin{tabularx}{\linewidth}{@{}>{\raggedright\arraybackslash}p{0.2\linewidth}>{\raggedright\arraybackslash}X>{\raggedright\arraybackslash}X@{}}
\toprule
\textbf{Variant} & \textbf{Change from the full method} & \textbf{Component tested} \\
\midrule
w/o novelty reward & Remove the novelty reward $r_{\mathrm{nov}}$ from $\rQ$ & Diversity of proposed questions \\
w/o learnability reward & Remove the learnability reward $r_{\mathrm{learn}}$ from $\rQ$ & Targeting of questions at the Solver's capability \\
w/o Questioner & Train the Solver only on canonical questions, without Questioner proposals & Self-proposed questions \\
w/o evidence score & Remove the evidence score $r_{\mathrm{evid}}$ from $\rS$ & Statement-level $F_1$ of evidence lines \\
w/o grounded score & Remove the grounded score $r_{\mathrm{grd}}$ from $\rS$ & Interval credit on matched lines \\
Answer reward only & Keep only the answer reward $r_{\mathrm{ans}}$ in $\rS$ & Both evidence terms \\
w/o rule selection & Keep the initial rules in every round; the curriculum still reallocates instances & Rule changes \\
\bottomrule
\end{tabularx}
\end{table}

\subsection{Training Value of Evolved Environments}
\label{app:task-value}

The comparison of Table~\ref{tab:env-source} in \S\ref{sec:analysis} fine-tunes the base Qwen3-VL-4B-Instruct checkpoint for one epoch on the questions stored in the bank at every round of one run, paired with the Solver responses recorded for them.
The two runs are the ablation that keeps the initial rules (w/o rule selection) and the full method, whose rules evolve through rule selection; they use the same curriculum and budget and differ only in rule selection.
The fine-tuned model does not select rules in this comparison, so differences between rows come from the data that rule selection produced.

\FloatBarrier
\section{Prompt Templates and Qualitative Rollout}
\label{app:prompts}

The Questioner prompt (Listing~\ref{lst:questioner}) provides the video frames, the scene description and active rules, the askable criteria, one bank question of the same task, and the Solver's pass rate per criterion; the listing shows its task statement and output format.
The Solver prompt (Listing~\ref{lst:solver}) specifies the format of evidence lines and answers.
Both prompts remain fixed across all experiments; bracketed markers show where task specifications and environment inputs are inserted.

\begin{lstlisting}[style=prompt,caption={Questioner prompt (task statement and output format).},label={lst:questioner}]
Write one multiple-choice question for one verified query below. Select a
query whose answer depends on when things happen: time-anchored observations,
their order, a duration, or the interval of an event.

[Rules on keeping the query, choosing admissible options, and the target
Solver success are inserted here.]

Before the proposal, find in the video the observations that determine the
selected query's answer and write them as located evidence lines, with no tag
around them and no text before them: one line per observation. The proposal's
grounding credit comes from these lines alone. Start each line with that
observation's time span in this video's seconds, read from the frame
timestamps, [[start,end]] or [[start,end],[start,end]], followed by one claim
stating the observation. A span copied from the query or the options earns no
credit. Then give the proposal between <answer> and </answer>, and write
nothing after </answer>.

[Scene and rules, verified queries with their distractor pools, current
feedback, and one prior example are inserted here.]

Return exactly:
[[START,END]] ONE LINE PER OBSERVATION THAT DETERMINES THE ANSWER
<answer>
criterion: QUERY_ID
question: QUESTION
options:
A. EXACT VERIFIED VALUE FROM THE SELECTED QUERY
B. DISTINCT DISTRACTOR FROM THAT QUERY
...
</answer>
\end{lstlisting}

\begin{lstlisting}[style=prompt,caption={Solver prompt (head and evidence-line rules).},label={lst:solver}]
Answer only the current question using the video.
Identify the observations needed to derive the answer: each relevant event,
or the visible states and operations being compared. Locate each observation
in the video before answering.
Each located line is checked on its own against the observations recorded
for this video: its claim must state one of them with the same object and
every identifying detail, and its interval must cover the time the video
shows it. A vague, merged or concluded claim matches none.
[Rules on naming subjects and choosing intervals are inserted here.]
Use as many evidence lines as the question needs. Each observation is
credited on one line; repeated copies of an observation and lines that state
no recorded observation lower the score. After the located lines, derive the
requested count, order, interval or state in plain lines without intervals,
then answer.
Start each line with its interval in this video's seconds, read from the
timestamps shown with the frames: [[START,END]], or [[START,END],[START,END]]
for separate occurrences.
[The answer instruction, worked examples, the question, its options, and the
video are inserted here.]
\end{lstlisting}

\noindent\textbf{Representative rollout example.}
Listing~\ref{lst:example} shows a Solver response on a Billiards question.
The question asks how the target ball's quadrant changes between the start and the end of the video, with six options, where option \texttt{F} is a shift from the upper-left to the lower-left quadrant.
The Solver cites two intervals corresponding to the initial and terminal observation frames, and its final line states the conclusion in the form the criterion is checked against.

\begin{lstlisting}[style=prompt,caption={A Solver response on a Billiards question.},label={lst:example}]
[[0.10, 0.60]] The pink ball is positioned in the upper-left quadrant at the start.
[[47.70, 48.00]] By the end, the pink ball is located in the lower-left quadrant, showing a movement from upper-left to lower-left.
<answer>F</answer>
\end{lstlisting}

\section{Limitations and Broader Impact}
\label{app:limitations}

\noindent\textbf{Limitations.}
Three constraints bound this study.
First, the instruction-tuned backbones inherit visual and linguistic abilities from pretraining, and the executable environments shape only the self-play stage that follows.
Second, the environments cover twelve hand-designed families of physical behavior with fixed task schemas and rule menus, and natural video has greater perceptual variability and commonsense breadth.
Third, supervision relies on deterministic rendering and exact temporal coordinates, and the evidence terms credit only statements that match the verification criteria.

\noindent\textbf{Broader impact.}
\method{} derives its training signal from executed programs, which reduces the need for scraped video and human temporal annotation.
Because every training video is rendered from synthetic programs, self-play uses no private, copyrighted, or personal video data.
As in other synthetic training methods, rule selection could be used to steer models toward narrow visual priors.
Rule configurations are versioned, so the curriculum of every run can be inspected.

%% file: sections/algorithm.tex
\begin{algorithm}[H]
\small
\caption{\method{} training loop}
\label{alg:tempo}
\begin{algorithmic}[1]
\REQUIRE policy $\pi_\theta$, frozen reference policy $\pi_{\mathrm{ref}}$, program library $\library$, empty bank $\bank$
\FOR{round $r=1,\ldots,R$ with playback scale $\lambda_r$}
 \STATE fix the allocation $\omega_{f,a}$ of round $r$ \hfill Eq.~\ref{eq:weight}
 \FOR{each step $t$}
  \STATE $\mathcal{W}\leftarrow\textsc{Batch}$: next batch of the round's allocation, screened and rendered
  \STATE sample proposals $q\sim\pi_{\theta_t}^{\mathrm{Q}}(\cdot\mid w)$ for each $w\in\mathcal{W}$; compute gate $g(q)$
  \STATE $\hat{\mathcal{Q}}\leftarrow$ gated proposals, at most one per instance, preselected by class and task rarity
  \STATE $\mathcal{Q}\leftarrow\{\bar q_w\}_{w\in\mathcal{W}}\cup\hat{\mathcal{Q}}\cup{}$ questions replayed from $\bank$
  \STATE sample $M$ rollouts $y\sim\pi_{\theta_t}^{\mathrm{S}}(\cdot\mid v,q)$ for each gated proposal and each $q\in\mathcal{Q}$; compute $p(q)$
  \STATE compute $\rQ$ for all proposals and $\rS$ for rollouts on $\mathcal{Q}$ \hfill Eqs.~\ref{eq:evidence_scores}--\ref{eq:learnability}
  \STATE record measured canonical questions and $\hat{\mathcal{Q}}$ in $\bank$; evict overrepresented task classes and families, then the oldest
  \STATE mean-center rewards within each group; omit groups whose rewards are all equal
  \STATE $\theta_{t+1}\leftarrow$ one GRPO update on Solver groups over $\mathcal{Q}$ and Questioner groups per instance
 \ENDFOR
 \IF{$r<R$}
  \STATE move each pair's level by its decayed pass rate in $\bank$
  \STATE for each family, the policy sets class weights in $w_{f,a}$ and keeps $u_f$ or selects a rule change from its menu
  \STATE the executor validates and previews each change; an invalid change keeps the previous program
 \ENDIF
\ENDFOR
\RETURN $\pi_\theta$
\end{algorithmic}
\end{algorithm}